\documentclass[letterpaper,conference,compsoc,10pt]{IEEEtran}
\IEEEoverridecommandlockouts

\usepackage{blindtext}
\usepackage{fancyhdr}

\usepackage[square, comma, sort&compress, numbers]{natbib}
\let\cite\citep
\setcitestyle{comma}
\setcitestyle{open={[},close={]},sep={,}}

\usepackage{amsmath,amssymb,amsfonts}
\usepackage{algorithmic}
\usepackage{graphicx}
\usepackage{textcomp}
\usepackage{setspace}
\usepackage{booktabs}
\usepackage{url}
\usepackage{enumitem}
\usepackage{mdframed}
\definecolor{lightgray}{RGB}{245, 245, 245}
\usepackage[table,xcdraw]{xcolor}

\usepackage{multirow}
\usepackage{caption}
\usepackage{balance}
\usepackage{subcaption}
\usepackage{array}
\usepackage{diagbox}
\usepackage[normalem]{ulem}
\useunder{\uline}{\ul}{}
\usepackage{siunitx} 

\usepackage{algorithm}
\usepackage{algorithmic}

\usepackage[colorlinks,
            linkcolor=red,
            anchorcolor=red,
            citecolor=green]{hyperref} %

\usepackage[capitalize]{cleveref}
\Crefname{section}{Sec.}{Secs.}
\Crefname{table}{Tab.}{Tabs.}
\Crefname{equation}{Eq.}{Eqs.}
\Crefname{figure}{Fig.}{Figs.}
\Crefname{lemma}{Lemma}{Lemmas}
\Crefname{theorem}{Theorem}{Theorems}
\Crefname{definition}{Definition}{Definitions}
\Crefname{hypothesis}{Hypothesis}{Hypothesises}

\def\ie{\textit{i.e.}}
\def\eg{\textit{e.g.}}

\newcommand{\vpara}[1]{\vspace{0.05in}\noindent \textbf{#1 }}

\usepackage[most]{tcolorbox}

\definecolor{boxcolor}{RGB}{230,240,250}

\newenvironment{remark}[1][]
  {
 \begin{tcolorbox}
 [%
    enhanced, 
    breakable,
    boxrule=0.5pt,
    arc=4pt,
    left=2pt,
    right=2pt,
    bottom=2pt,
    top=2pt,
    rounded corners
    ]{}
  \textbf{#1.}
  \small \itshape
  }
  {
\end{tcolorbox}
}

\renewcommand{\figurename}{Figure} 

\newcommand{\myparatight}[1]{\smallskip\noindent{\bf {#1}:}~}

\begin{document}
\pagenumbering{arabic}

\title{Securely Fine-tuning Pre-trained Encoders Against Adversarial Examples}


\author{
\IEEEauthorblockN{ 
Ziqi Zhou\textsuperscript{1,2,3}\IEEEauthorrefmark{1},
Minghui Li\IEEEauthorrefmark{2},
Wei Liu\textsuperscript{1,2,4,5}\IEEEauthorrefmark{3},
Shengshan Hu\textsuperscript{1,2,4,5}\IEEEauthorrefmark{3}, 
Yechao Zhang\textsuperscript{1,2,4,5}\IEEEauthorrefmark{3} \\
Wei Wan\textsuperscript{1,2,4,5}\IEEEauthorrefmark{3},
Lulu Xue\textsuperscript{1,2,4,5}\IEEEauthorrefmark{3},
Leo Yu Zhang\IEEEauthorrefmark{4},
Dezhong Yao\textsuperscript{1,2,3}\IEEEauthorrefmark{1}, 
and Hai Jin\textsuperscript{1,2,3}\IEEEauthorrefmark{1}
}

\textsuperscript{1}National Engineering Research Center for Big Data Technology and System\\
\textsuperscript{2}Services Computing Technology and System Lab \
\textsuperscript{3}Cluster and Grid Computing Lab 
\\
\textsuperscript{4}Hubei Engineering Research Center on Big Data Security \ \textsuperscript{5}Hubei Key Laboratory of Distributed System Security \\

\IEEEauthorblockA{
\IEEEauthorrefmark{1}
School of Computer Science and Technology, 
Huazhong University of Science and Technology \\
}

\IEEEauthorblockA{
\IEEEauthorrefmark{2}
School of Software Engineering, 
Huazhong University of Science and Technology \
}

\IEEEauthorblockA{
\IEEEauthorrefmark{3}
 School of Cyber Science and Engineering,
 Huazhong University of Science and Technology\\
}

\IEEEauthorblockA{
\IEEEauthorrefmark{4}
School of Information and Communication Technology,
Griffith University\
}

{\tt \footnotesize \{zhouziqi, minghuili, weiliu73, hushengshan, ycz, wanwei\_0303, lluxue, dyao, hjin\}@hust.edu.cn, 
}
\\
{\tt \footnotesize leo.zhang@griffith.edu.au
}

}
\maketitle

\begin{abstract}
With the evolution of self-supervised learning, the pre-training paradigm has emerged as a predominant solution within the deep learning landscape. Model providers furnish pre-trained encoders designed to function as versatile feature extractors, enabling downstream users to harness the benefits of expansive models with minimal effort through fine-tuning.
Nevertheless, recent works have exposed a vulnerability in pre-trained encoders, highlighting their susceptibility to downstream-agnostic adversarial examples (DAEs) meticulously crafted by attackers. The lingering question pertains to the feasibility of fortifying the robustness of downstream models against DAEs, particularly in scenarios where the pre-trained encoders are publicly accessible to the attackers.

In this paper, we initially delve into existing defensive mechanisms against adversarial examples within the pre-training paradigm. Our findings reveal that the failure of current defenses stems from the domain shift between pre-training data and downstream tasks, as well as the sensitivity of encoder parameters. In response to these challenges, we propose \textit{\underline{\textbf{G}}enetic \underline{\textbf{E}}volution-\underline{\textbf{N}}urtured \underline{\textbf{A}}dversarial \underline{\textbf{F}}ine-tuning (Gen-AF)}, a two-stage adversarial fine-tuning approach aimed at enhancing the robustness of downstream models.
Gen-AF employs a genetic-directed dual-track adversarial fine-tuning strategy in its first stage to effectively inherit the pre-trained encoder. This involves optimizing the pre-trained encoder and classifier separately while incorporating genetic regularization to preserve the model's topology. 
In the second stage, Gen-AF assesses the robust sensitivity of each layer and creates a dictionary, based on which the top-k robust redundant layers are selected  with the remaining layers held fixed.
Upon this foundation, we conduct evolutionary adaptability fine-tuning to further enhance the model's generalizability.
Our extensive experiments, conducted across ten self-supervised training methods and six datasets, demonstrate that Gen-AF attains high testing accuracy and robust testing accuracy against state-of-the-art DAEs. 
\end{abstract}
    
\section{Introduction} 
\label{sec:intro}
With triumphs of deep learning, researchers are dedicated to training models with strong performance to address the multifaceted challenges in the real world.
However, constructing a capable model requires substantial labeled datasets and expensive computational resources, which significantly hinders their applications.
Recently, the emergence of large pre-trained encoder (\eg, GPT~\cite{brown2020language}, SimCLR~\cite{chen2020simple}, CLIP~\cite{radford2021learning}) has significantly alleviated resource constraints, which are trained  by model providers (\eg, Google, Meta, and OpenAI) through \textit{self-supervised learning} (SSL) methods.
Fine-tuning the pre-trained encoder, which enjoys powerful feature extraction prowess and knowledge transfer aptitude,  is emerging as a new  deep learning paradigm. 
For example, the CLIP provided by OpenAI has facilitated the flourishing of downstream tasks like Semantic Segmentation~\cite{zhou2023zegclip}, Video Processing~\cite{rasheed2023fine}, 3D Point Cloud Classification~\cite{ huang2023clip2point}.

However, recent works have uncovered the potential security risks of pre-trained encoders~\cite{jia2022badencoder, saha2022backdoor, liu2022poisonedencoder,ban2022pre,zhou2023advencoder}. 
PoisonedEncoder~\cite{liu2022poisonedencoder} injects crafted poisoned examples into the unlabeled pre-training data such that  the target downstream classifiers inheriting the poisoned encoder will misclassify certain samples into target class.
In a similar vein, BadEncoder~\cite{jia2022badencoder} embeds a backdoor by fine-tuning a pre-trained encoder and subsequently disseminates the backdoored pre-trained encoder on third-party platforms (\eg, GitHub, hugging face).
Any downstream model utilizing this encoder can be activated through a carefully designed trigger by an attacker  for malicious purposes.
While these backdoor attacks appear concerning, defense strategies are straightforward and easy-to-deploy, such as cleaning  pre-training datasets or avoiding downloading from untrusted sources.

 \begin{figure}[!t]
    \centering
    \includegraphics[scale=0.54]{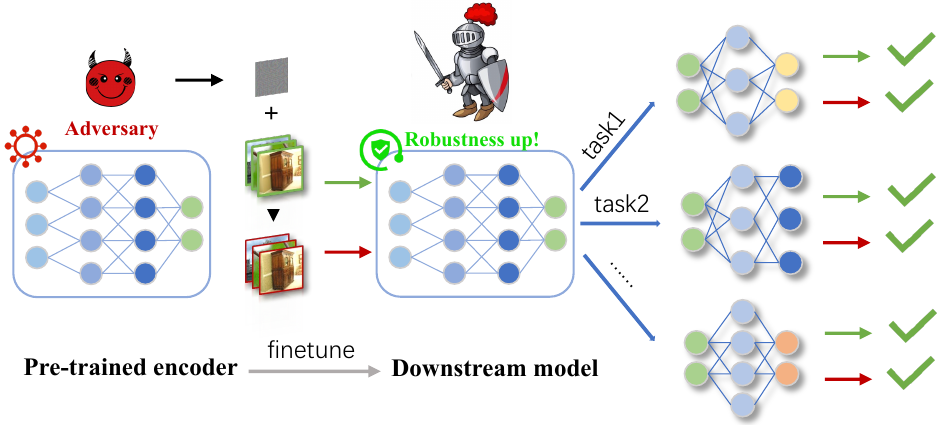}
    \caption{Overview of downstream users mitigating DAEs.}
    \label{fig:demo}
\end{figure}

Unlike backdoor attacks occurring in the training phase, adversarial attacks~\cite{goodfellow2014explaining, carlini2017towards, zhang2024whydoes} stemming from the model's inherent vulnerabilities~\cite{goodfellow2014explaining} occur during the inference phase and are more threatening.
Latest works~\cite{ban2022pre, zhou2023advencoder, zhou2023advclip} have successfully utilized a publicly available pre-trained encoder to craft \textit{downstream-agnostic adversarial examples}
(DAEs) to fool   downstream tasks.
Specifically, PAP~\cite{ban2022pre} leverages the stability of shallow layers in a pre-trained encoder to design pre-trained perturbations for attacking downstream models after fine-tuning.
AdvEncoder~\cite{zhou2023advencoder} utilizes a frequency-based generative framework to craft DAEs.
These works demonstrate that, despite  the lack of knowledge  about the pre-training dataset and downstream tasks, attackers can still craft highly effective adversarial examples  to compromise all the downstream tasks based on publicly available pre-trained encoders. This poses an intriguing problem:
\begin{quote}
    \emph{Is it feasible to conduct a secure fine-tuning of a pre-trained encoder to develop downstream models with resilience against DAEs, especially when the attacker possesses knowledge of the pre-trained encoder?}  
\end{quote}
Notably, there are many solutions in effectively detecting adversarial examples like~\cite{tao2018attacks, yin2019gat, liang2018detecting}. They are orthogonal to our design goals and can be easily integrated.  

In this study, we start by examining existing defenses against adversarial examples to gauge their effectiveness in the pre-training paradigm. Our goal is to uncover the challenges in dealing with  DAEs, as illustrated in Figure \ref{fig:demo}. Through extensive experiments on four kinds of defense methods, we find that none of them effectively counter recently proposed DAEs.
We owe this to  two key characteristics of  pre-training: 1) The domain shift, caused by the distinct training data types between pre-training and downstream fine-tuning,  makes it harder for the model to differentiate between benign examples and DAEs due to reduced diversity in the feature space; 2) The high sensitivity of pre-trained encoder, caused by  catastrophic knowledge forgetting (\ie, loss of original feature extraction capability), significantly constrains the ability of the defender as  achieving optimal model generalization is more desired.  This puts us into a ``pre-training dilemma" situation. 

To address the aforementioned challenges, we present the inaugural downstream method, \textit{\underline{\textbf{G}}enetic \underline{\textbf{E}}volution-\underline{\textbf{N}}urtured \underline{\textbf{A}}dversarial \underline{\textbf{F}}ine-tuning (Gen-AF)}. Our objective is to enhance the robustness of downstream models while preserving the inherent generalization capabilities of the pre-trained encoder. Gen-AF operates through a two-stage adversarial fine-tuning approach, comprising genetic-driven dual-track adversarial fine-tuning and evolutionary adaptability fine-tuning.
In the first stage, to seamlessly inherit the pre-trained encoder,
we introduce a bilevel-optimizer collaborative strategy. This strategy optimizes the parameters of the pre-trained encoder and the classifier separately, assigning a minimal learning rate for the pre-trained encoder and a standard learning rate for the classifier. Additionally, to counteract the potential reduction in model generalization during adversarial training, we implement genetic regularization. This technique maintains the relative positional relationships of natural samples within the representation space, thereby preserving the model's inherent generalization capacity.

Moving to the second stage, our aim is to retain the robust-sensitive layers trained adversarially while fine-tuning the robust-redundant layers. We construct a sensitivity dictionary for each network layer of the downstream model and select the top-k layers with the lowest robustness, keeping the remaining layers fixed. Subsequently, we undertake standard fine-tuning training for the selected layers to enhance the overall generalization of the downstream model.
We verify the  performance of Gen-AF against five SOTA universal adversarial attacks designed for pre-trained encoders, across ten popular SSL training methods, two pre-training datasets, and six downstream datasets.
The results demonstrate that our method can effectively defend against  DAEs, 
achieving a well-balanced trade-off between  robustness and generalization.

Our main contributions are summarized as follows:
\begin{itemize}
\item We extensively investigate existing defenses in the pre-trained paradigm and offer a comprehensive understanding in mitigating DAEs.
\item We design the first genetic evolution-nurtured adversarial fine-tuning to bolster the robustness of downstream models while simultaneously maintaining the generalization ability inheriting from the pre-trained encoder. 
\item Our extensive experiments on ten self-supervised training methods and six datasets show that  Gen-AF achieves high testing accuracy and robust testing accuracy against state-of-the-art DAEs.  
The results also demonstrate that Gen-AF can defend against backdoor attacks targeting pre-trained encoders without any modifications.
\end{itemize}

\section{Preliminaries}\label{sec:motivation}

\subsection{Threat Model}\label{sec:threat_model}
We evaluate the security risks of pre-trained models from the perspectives of  attackers and downstream task undertakers (called defenders hereinafter).

\myparatight{Attacker's knowledge and capabilities} 
We assume that attackers have various approaches to access pre-trained encoder but lacking knowledge of downstream tasks. 
Depending on their knowledge of the pre-training process, attackers fall into the following three categories: 
\begin{itemize}
    \item {\bf Full upstream-knowledge attacker.} 
    The attackers, such as model providers or their affiliates, can access to both the pre-trained encoder and the pre-training dataset, based on which adversarial examples could be crafted. 
    \item {\bf Partial upstream-knowledge attacker.} 
    The attackers, such as third-party malicious attackers, can only access the pre-trained encoder but cannot obtain the pre-training dataset. 
    Despite this limitation, they can use the pre-trained encoder with an unrelated surrogate dataset to fabricate adversarial examples.   
    \item {\bf Transfer-based black-box attacker.} The attackers lack information about the pre-trained encoder and the dataset. They employ transfer-based methods to attack downstream models.
\end{itemize}

\myparatight{Attacker's goals}
Following~\cite{ban2022pre, zhou2023advencoder}, 
we assume that adversaries are inclined to exploit publicly accessible pre-trained encoders, obtained through means such as purchase or direct download from publicly available platforms. The goal is to create adversarial examples targeting downstream models based on these encoders to launch \textit{non-targeted} attacks, given the adversaries' limited knowledge of downstream specifics. 
Their goals encompass three key features: 

\begin{itemize}
    \item {\bf Universality.} 
    The adversarial examples should effectively compromise any downstream models utilizing the same pre-trained encoder, irrespective of the specific downstream tasks involved.
    \item {\bf Effectiveness.} 
    The adversarial examples can undermine  the functionality of downstream models  even after undergoing fine-tuning and any conceivable defensive measures.
    \item {\bf Stealthiness.} Perturbations added to   adversarial examples should be imperceptible to human beings so as not to be easily detected.
\end{itemize}

\begin{figure*}[!t]  
  \centering
         \subcaptionbox{IP-TA \label{2_a}}{\includegraphics[width=0.173\textwidth]{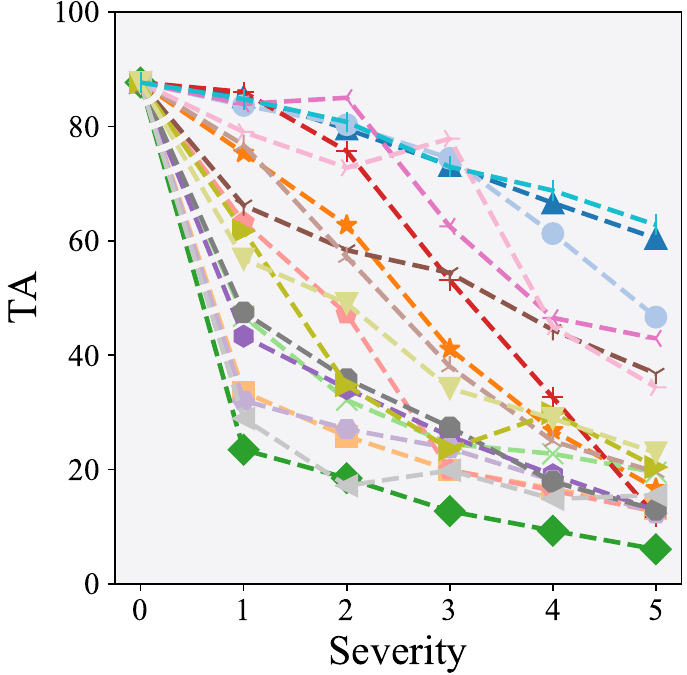}} 
            \subcaptionbox{IP-ASR-A \label{2_b}}{\includegraphics[width=0.172\textwidth]{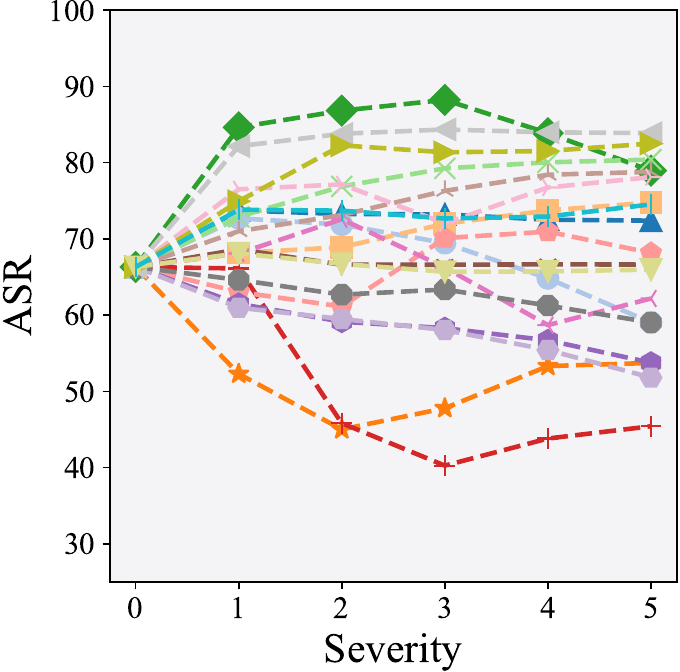}}
     \subcaptionbox{IP-RA-A \label{2_c}}{\includegraphics[width=0.168\textwidth]{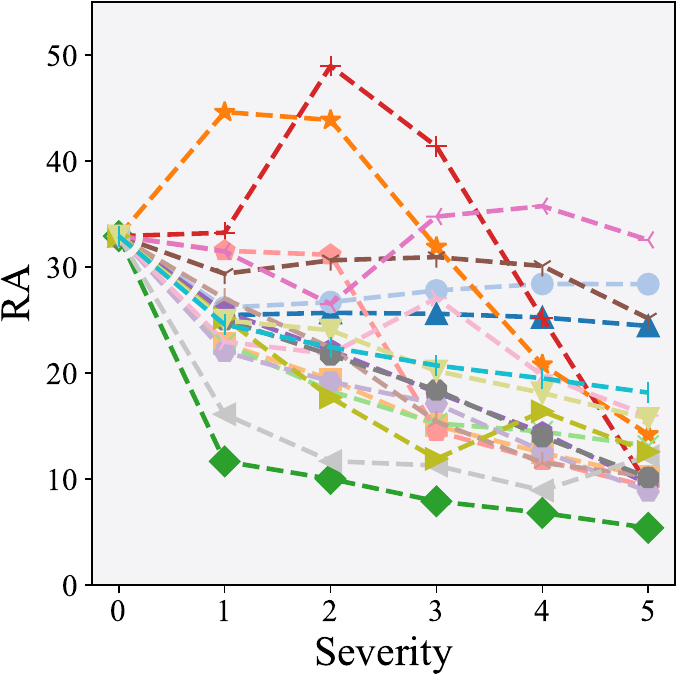}}
        \subcaptionbox{IP-ASR-P \label{2_d}}{\includegraphics[width=0.171\textwidth]{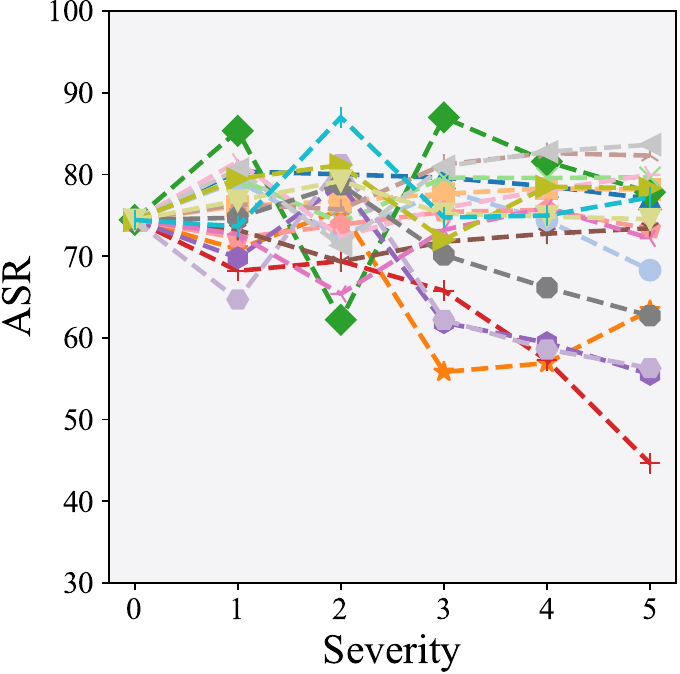}}
    \subcaptionbox{IP-RA-P \label{2_e}}{\includegraphics[width=0.17\textwidth]{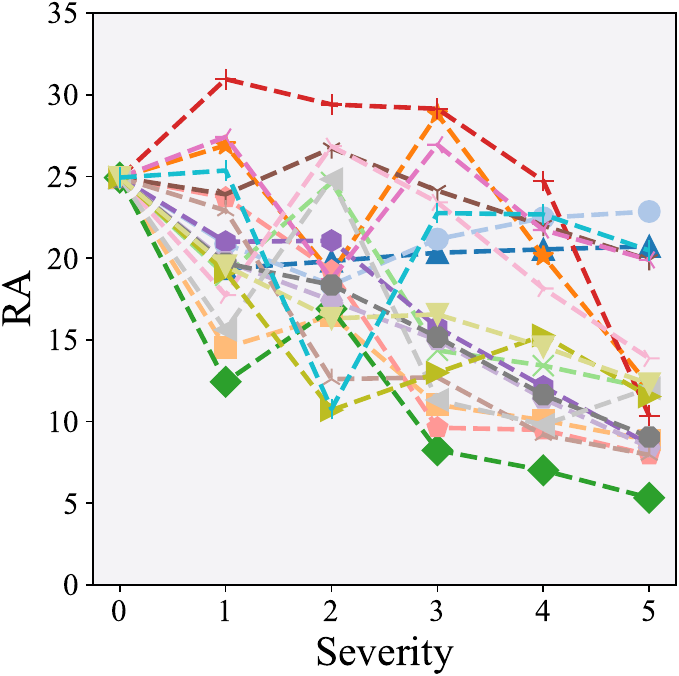}}
    \hspace{0.38cm}
    \includegraphics[width=0.07\textwidth]{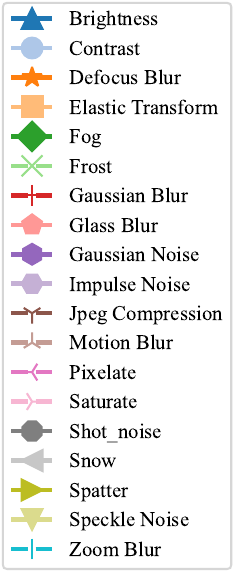}
    \subcaptionbox{PR-A \label{2_f}}{\includegraphics[width=0.16\textwidth]{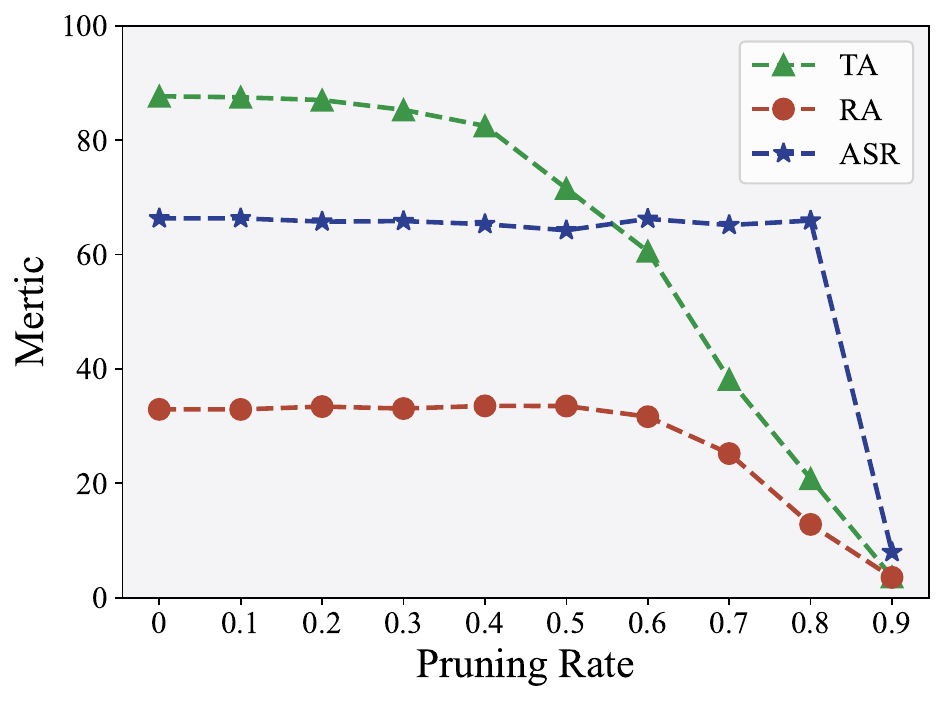}}
    \subcaptionbox{PR-P \label{2_g}}{\includegraphics[width=0.16\textwidth]{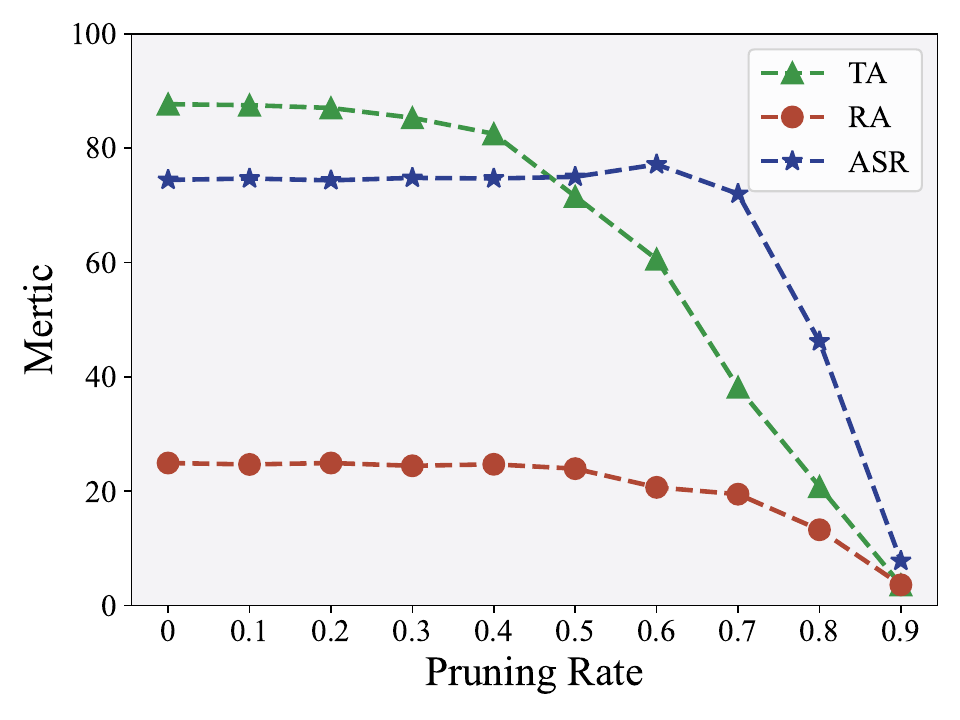}}
    \subcaptionbox{Dist-I \label{2_h}}{\includegraphics[width=0.16\textwidth]{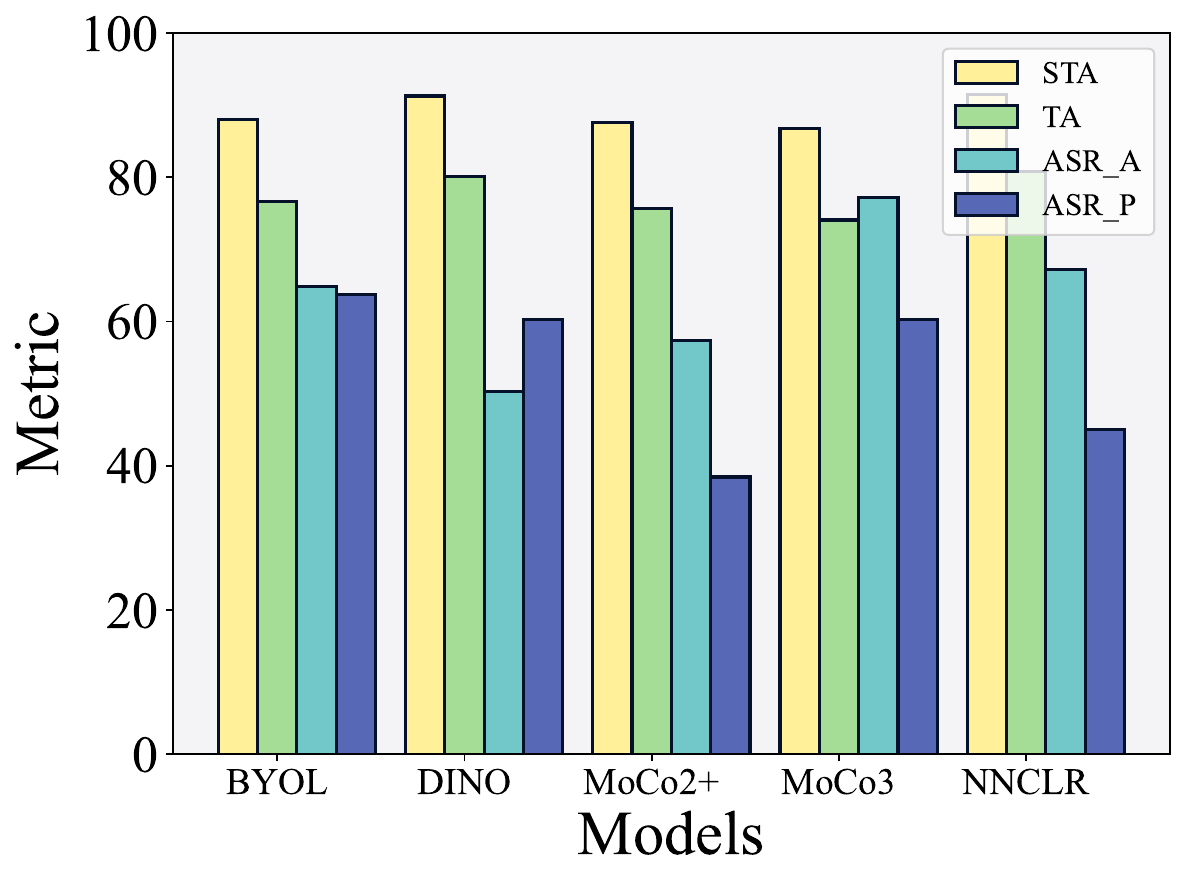}}
    \subcaptionbox{Dist-II \label{2_i}}{\includegraphics[width=0.16\textwidth]{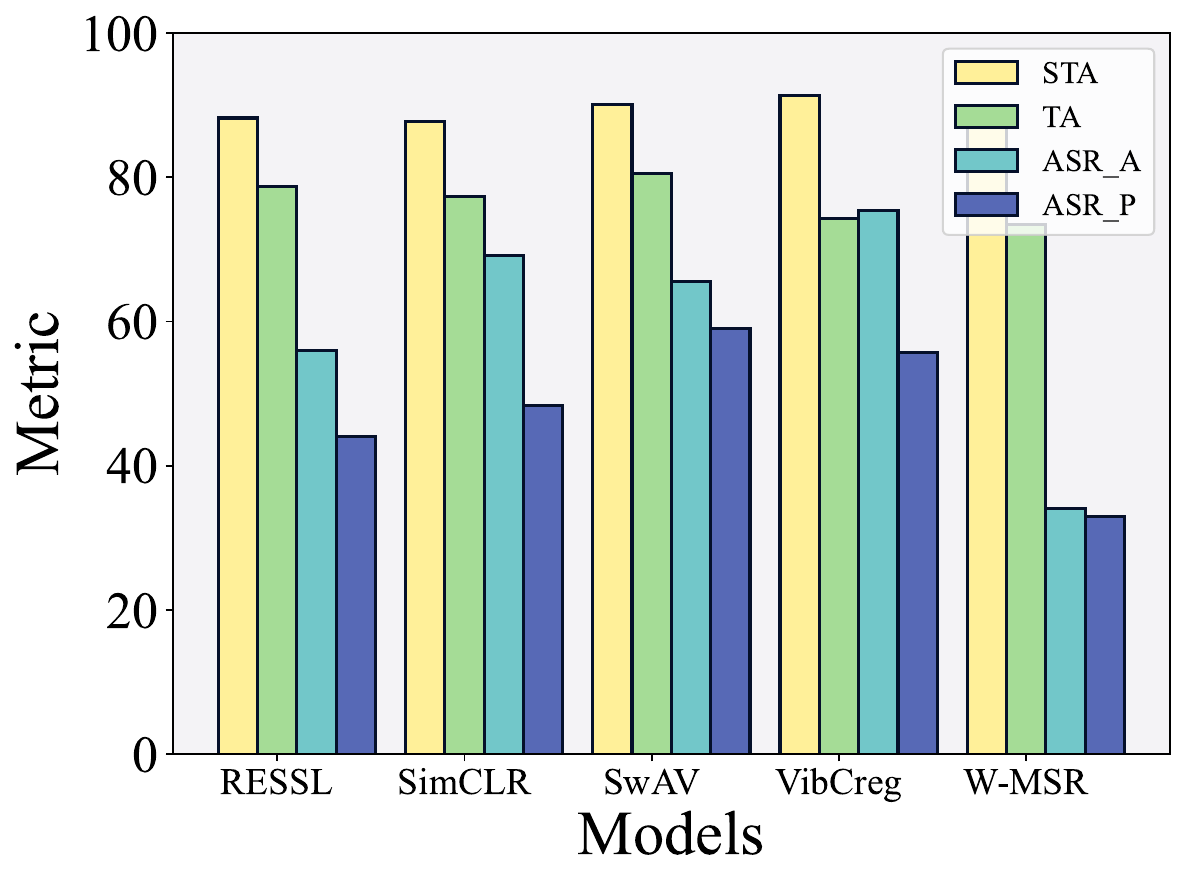}}
  \subcaptionbox{AT-I \label{2_j}}{\includegraphics[width=0.158\textwidth]{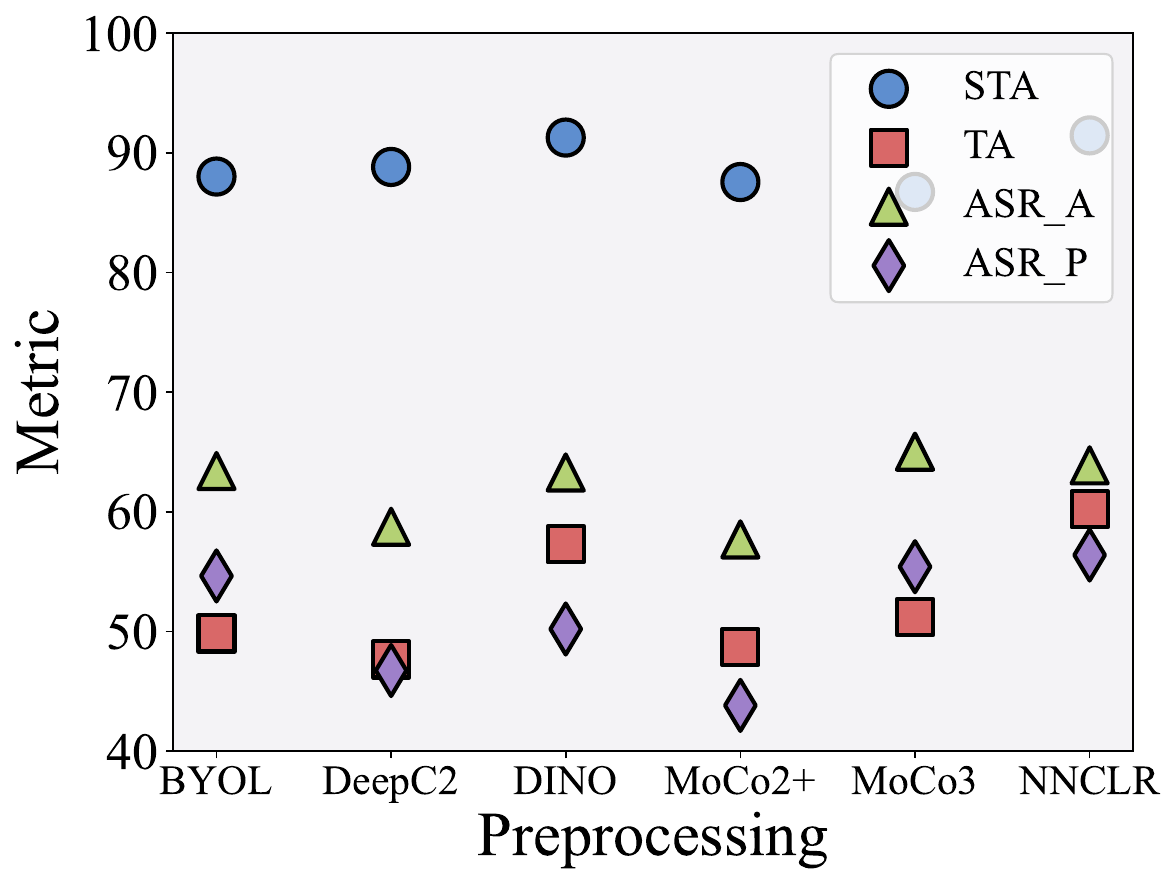}}
    \subcaptionbox{AT-II \label{2_k}}{\includegraphics[width=0.16\textwidth]{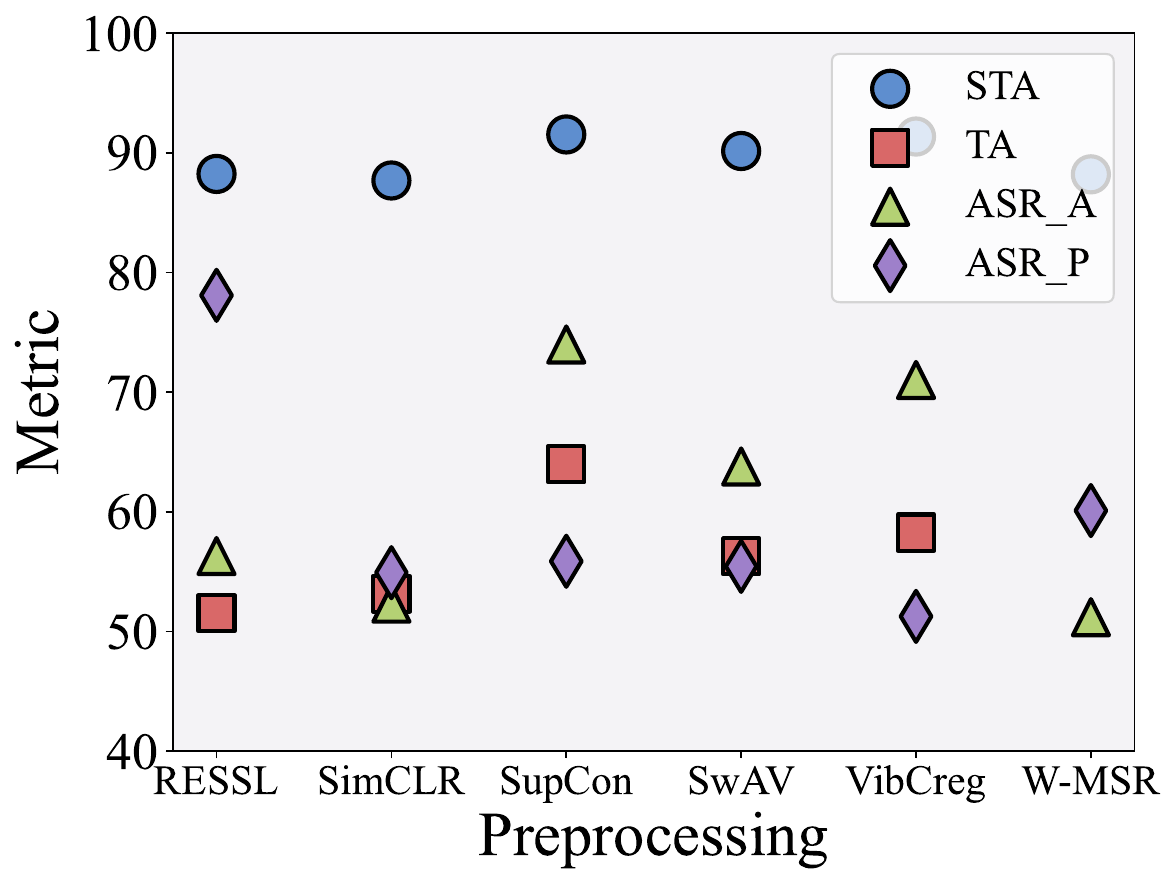}}
      \caption{Experimental results of various defense methods in mitigating downstream-agnostic adversarial examples. ASR-A and  RA-A represent the attack success rate and robust accuracy of adversarial examples created using AdvEncoder. ASR-P and RA-P denote the same results for PAP. Figures (a) - (e) for image preprocessing (IP), (f) - (g) parameter pruning (PR), (h) - (i) model distillation (Dist), (j) - (k) adversarial training (AT).}
       \label{fig:defense}
        \vspace{-0.4cm}
\end{figure*}

\myparatight{Defender's knowledge and capabilities} 
Defenders are not in charge of the model's pre-training procedure and lack knowledge about pre-training process, but they retain authority over the fine-tuning process of the downstream model, including data processing and training.

\myparatight{Defender's goals}
We consider that the defender has limited computing resources and aims to fine-tune pre-trained encoders in an efficient way to complete  downstream tasks, while concurrently mitigating the adversarial attacks.
Specifically, the defender has the following three goals: 

\begin{itemize}
    \item {\bf Generalization.} 
   The goal entails ensuring that models built on pre-trained encoders achieve high classification accuracy on downstream datasets, even with defensive measures in place.
    \item {\bf Robustness.} 
    The goal is for downstream models to effectively mitigation adversarial examples, especially those created with pre-trained encoders.
    \item {\bf Resource efficiency.} 
    The goal refers to defenders being limited to using only the data and models from the original task for their defensive  strategies.    
\end{itemize}

\subsection{Investigation into Existing Defenses}\label{sec:investigating}

In this section, we delve into the effectiveness of existing  defense strategies against adversarial examples in downstream task scenarios.
Our analysis focuses on identifying and understanding their constraints, thereby shedding light on the inherent challenges associated with the defenses in the pre-training paradigm.  
We examine existing defenses from two perspectives: sample level and model level. We evaluate their effectiveness against the  latest downstream-agnostic adversarial examples (DAEs)  (\ie, AdvEncoder~\cite{zhou2023advencoder} and PAP~\cite{ban2022pre}).

For sample-level strategies, we explore various data preprocessing~\cite{guo2017countering} methods such as noise addition, blurring, and compression to mitigate  adversarial examples. 
At the model level, we focus on model distillation~\cite{papernot2016distillation}, parameter pruning~\cite{zhu2017prune}, and adversarial training~\cite{ szegedy2013intriguing, carlini2017towards} to enhance model robustness.
For fairness, we evaluate  these approaches under the same partial upstream-knowledge attacker scenario where the attacker crafts DAEs using CIFAR10~\cite{krizhevsky2009learning} as the surrogate dataset on the pre-trained ResNet18 ImageNet~\cite{russakovsky2015ImageNet} encoder, and the GTSRB~\cite{stallkamp2012man} classification serves as the downstream task.
We use \textit{Testing Accuracy} (TA) and \textit{Robust Testing Accuracy} (RA) to evaluate the accuracy of the models to classify benign and adversarial examples, and \textit{Attack Success Rate} (ASR)  to evaluate the ability of  adversarial examples, respectively.
More details are described in \cref{sec:experiment_setting}.

\noindent\textbf{Solution I: Input preprocessing.} 
Preprocessing inputs with various techniques has proven effective against adversarial attacks by removing perturbations~\cite{guo2017countering, nie2022diffusion}. 
Recently, diffusion models  can serve as a purifier to defend against adversarial examples on the data domain of is training dataset~\cite{nie2022diffusion}.
However, training such models is extremely resource-intensive, cannot suffice the resource efficiency goal of a downstream defender.
Hence, we shift our focus to universal image defense methods that directly process images without requiring additional models and data.
As illustrated in \cref{fig:defense}, we utilize  \textit{nineteen}  transformations (\ie, corruptions) over downstream images, which includes: 
 \emph{(1) Noise-based:} Gaussian Noise (GN) , Shot Noise (SoN), Impulse Noise (IN), and Speckle Noise (SN). \emph{(2) Blur-based:} Defocus Blur (DB), Glass Blur (GB), Motion Blur (MB), Zoom Blur (ZB), and Gaussian Blur (GB). \emph{(3) Weather-based:} Snow (SW), Frost (FT), and Fog (FG).
\emph{(4) Quality Adjustments:} Brightness (BR), Contrast (CT), and Saturate (ST).
\emph{(5) Compression and Pixel Transformation:} JPEG Compression (JC),
Pixelate (PL). \emph{(6) Special Processing:} Elastic Transform (ET), Spatter (SP).
For each corruption, we set varying severity levels ranging from $0$ to $5$, where higher values indicate greater image degradation and more information loss. ``$0$'' represents images are unprocessed. 

From~\cref{2_b}, we observe that even with nineteen corruption schemes applied at the highest severity level of $5$, the lowest ASR for AdvEncoder remains above $40\%$. 
Similarly, in \cref{2_d}, at the highest severity level of $5$, except for Gaussian Blur, the lowest ASR for PAP remains above $55\%$.
It is noteworthy that some image preprocessing methods, while effective against adversarial examples, will notably reduce the model's generalization, even at lower severity levels, as shown in~\cref{2_a}.
For example, Fog and Snow
cause an average decline of over $50\%$ in TA  at the severity level of $1$.
Increased severity markedly decreases accuracy, as seen in the drop in both TAs in~\cref{2_a}, and RAs in~\cref{2_c} and~\cref{2_e} for severities 3 to 5. These findings underscore the impracticality of using input preprocessing methods to defend against adversarial examples in the pre-training paradigm.

\begin{remark}[Remark I]
Due to the inseparability of adversarial noise and the image, input preprocessing-based methods, while disrupting noise, also compromise the original image information, leading to significant accuracy loss. This is because the downstream task usually has different data domain and the fine-tuned model may not fit it well in a limited training steps, such that the diversity between benign examples and DAEs in the feature space becomes smaller.  
Moreover, such methods struggle to completely eliminate adversarial noise, thus limiting their effectiveness against adversarial examples. Consequently, input preprocessing-based schemes fall short in meeting the defender's goals for generalization and robustness.
\end{remark}

\noindent\textbf{Solution II: Adversarial Training.} 
Adversarial training~\cite{szegedy2013intriguing, carlini2017towards,zhang2019theoretically}, widely employed as a defense mechanism, enhances model robustness fundamentally by incorporating adversarial noise into the training dataset.
Compared to standard training, adversarial training exhibits increased complexity and presents a well-acknowledged challenge in striking a balance between generalization and robustness.
Aligned with the downstream fine-tuning approaches, there are two adversarial training strategie, adversarial training of only the linear layer and adversarial training of the entire model that contains the pre-trained encoder.
We choose TRADES~\cite{zhang2019theoretically}, the most popular adversarial training approach known for its theoretical generalization preservation, to study the effectiveness of the above adversarial training strategies in a pre-training scenario.

\emph{(1) Adversarial Training on Linear Layer:}
We conduct adversarial training on the downstream classifier using the GTSRB dataset, and as shown in \cref{2_j} and \cref{2_k}.
There is a noticeable decline in model accuracy (TA) compared to standard training (STA).
Training only the linear layers for adversarial robustness is limited, as evidenced by AdvEncoder and PAP still maintaining an average ASR value of over $50\%$.
This is due to the majority of the model's critical networks being frozen during training, 
making attacking the downstream models an quasi-white-box setup for the attacker.
Additionally, the robustness enhancement provided by shallow linear layers alone is limited.
\emph{(2) Adversarial Training on Entire Model:}
The primary issue  of adversarial training on the entire model is that  fine-tuning all the parameters in transfer learning requires a significant amount of training to strike the right balance between adapting the entire model to the new task and retaining the previously learned knowledge.
Moreover, current adversarial training methods also generally suffer from additional loss in model generalization. 
In \cref{sec:compare}, we employ TRADES for adversarial training, and the results indicate that while robustness is somewhat improved in the trained models, a notable decline in generalization is observed. 
This clearly contradicts the defender's generalizability goal.

\begin{remark}[Remark II]
The limitations of existing adversarial training in the ``pre-training'' paradigm manifest in the fact that adversarial training  requires optimizing sufficient amount of model parameters to enhance robustness, but this will also compromise the powerful feature extraction capabilities inherent in the original pre-trained parameters, which the model's generalizability depends on. This ``pre-training dilemma"  also makes existing adversarial training methods fail to work in the the pre-training paradigm,  as they predominantly focus on augmenting model performance from scratch and cannot adequately inherit the stability of   pre-trained encoder parameters.

\end{remark}

\noindent\textbf{Solution III: Parameter Pruning.} 
Parameter pruning~\cite{zhu2017prune}, as a prevalent downstream adaption operation,   involves reducing the complexity of a neural network by trimming non-critical parameters,  which makes it less sensitive to minor perturbations. 
We also experimentally show its effectiveness in defending against DAEs. 
As illustrated in \cref{2_f} and  \cref{2_g}, we prune the pre-trained encoders (based on SimCLR) at various pruning rates ranging from $0$ to $0.9$. The effectiveness of adversarial attacks only marginally decreases when at least $60\%$ of the model's parameters are pruned, by which point the model's accuracy has already dropped by nearly $30\%$, as shown in \cref{2_f}. 
Even with a pruning rate of $80\%$, where the model is substantially compromised, adversarial examples still maintain a high ASR.
The limitation of parameter pruning lies in its lack of customized design for adversarial samples, rendering it ineffective against DAEs.

\noindent\textbf{Solution IV: Model Distillation.} 
Model Distillation~\cite{papernot2016distillation} is a common downstream model compression technique that transfers knowledge from a complex pre-trained model (\ie, teacher model) to a more efficient model (\ie, student model). 
Existing works~\cite{papernot2016distillation} demonstrate that training a student model to imitate the teacher model's probabilistic output leads to smoother probability distributions. This smoothing effect makes the model less sensitive to minor input perturbations, thereby could possibly be used to defend against adversarial examples.
We use  the GTSRB dataset to distill ten SSL encoders, which are pre-trained on ImageNet, resulting in new distilled encoders. We then build downstream models on  GTSRB dataset to test the robustness of the distilled models  against AdvEncoder and PAP.
As shown in \cref{2_h} and \cref{2_i}, the distilled pre-trained encoders exhibit a noticeable decline in generalization performance with an average ASR still exceeding $50\%$.
This can be attributed to the loss of pre-trained knowledge due to variations in downstream data domains, and the strong cross-domain transferability inherent to DAEs.

\subsection{Key Challenges and Intuitions}
\label{sec:initution}
The success of DAEs depends on the stability of the pre-trained encoder's parameters during fine-tuning, allowing attackers to easily deceive downstream models in a quasi-white-box scenario. 
Inspired by biological evolution, we conceptualize the process of constructing downstream models as a question of how the fine-tuned models inherit the feature extraction capabilities (genes) from their original parameters of the pre-trained models in a new data domain, while simultaneously enhancing their generalizability and robustness (evolution). 
Our intuition is to strengthen the robustness of the pre-trained encoder through adversarial training.
However, the key of the pre-training paradigm lies in the well-trained model parameters, and any inappropriate alteration can lead to the collapse of its originally powerful feature extraction capabilities.
Therefore, the difficulty of using adversarial training to defend against DAEs stems from the aforementioned ``pre-training dilemma", which contains the following challenges:

\textbf{Challenge I: Resolving parameter conflicts of pre-trained encoder during fine-tuning.}
Compared to traditional adversarial training from scratch, our challenge lies in training the encoder to enhance its robustness while ensuring minimal parameter changes to prevent catastrophic forgetting of pre-trained knowledge. 
Existing studies~\cite{rice2020overfitting,chen2020robust,mi2023topology} indicate that adversarial training alters the distribution of benign examples in the feature space, resulting in the loss of generalizability. 
Standard training results in distinct intra- and inter-class relationships among benign examples, with same-class samples being closer and different-class samples further apart. 
In contrast, adversarial examples diverge from their original classes in the feature space. 
After adversarial training, although the distance between benign and adversarial examples reduces, it alters their original positional relationships, leading to the misclassification of benign examples.
Hence, we aim to constrain the divergence between adversarial and original benign examples while preserving the existing feature distribution of benign examples. This maintains the model's high accuracy brought about by the established feature boundaries.
Specifically, we construct a graph to capture the topological relationship (\ie, the inter-class and intra-class relationships of samples in the feature space). By reducing the disparity in the feature maps between adversarial and benign examples, we aim to enhance the stability of the model's feature space.
Considering the asynchrony in the fine-tuning of pre-trained encoders and linear layers, we further design a bilevel-optimizer collaborative strategy, replacing the traditional single-optimizer paradigm, setting a smaller learning rate for the pre-trained encoder and a standard learning rate for the linear layers. 
The experimental results in \cref{sec:ablation} prove the necessity of the bilevel-optimizer collaborative strategy.

\textbf{Challenge II: Balancing trade-off between generalization and robustness.}
Despite our refined fine-tuning of pre-trained encoder parameters to enhance model robustness, we still confront the trade-off between generalizability and robustness. 
Diverging from traditional adversarial training methods that seek a dynamic balance in a single training process, we adopt a divide-and-conquer approach, integrating adversarial and standard training. Initially, we employ our adversarial training to obtain a robustness-enhanced preliminary model. 
Subsequently, we assess the contribution of each network layer to robustness, \ie, their sensitivity to adversarial noise. 
Finally, we select the top-k robustness-insensitive layers and freeze the others for standard training, thereby improving generalization without compromising the existing robustness.

 \begin{figure*}[!t]
    \centering
    \includegraphics[scale=0.52]{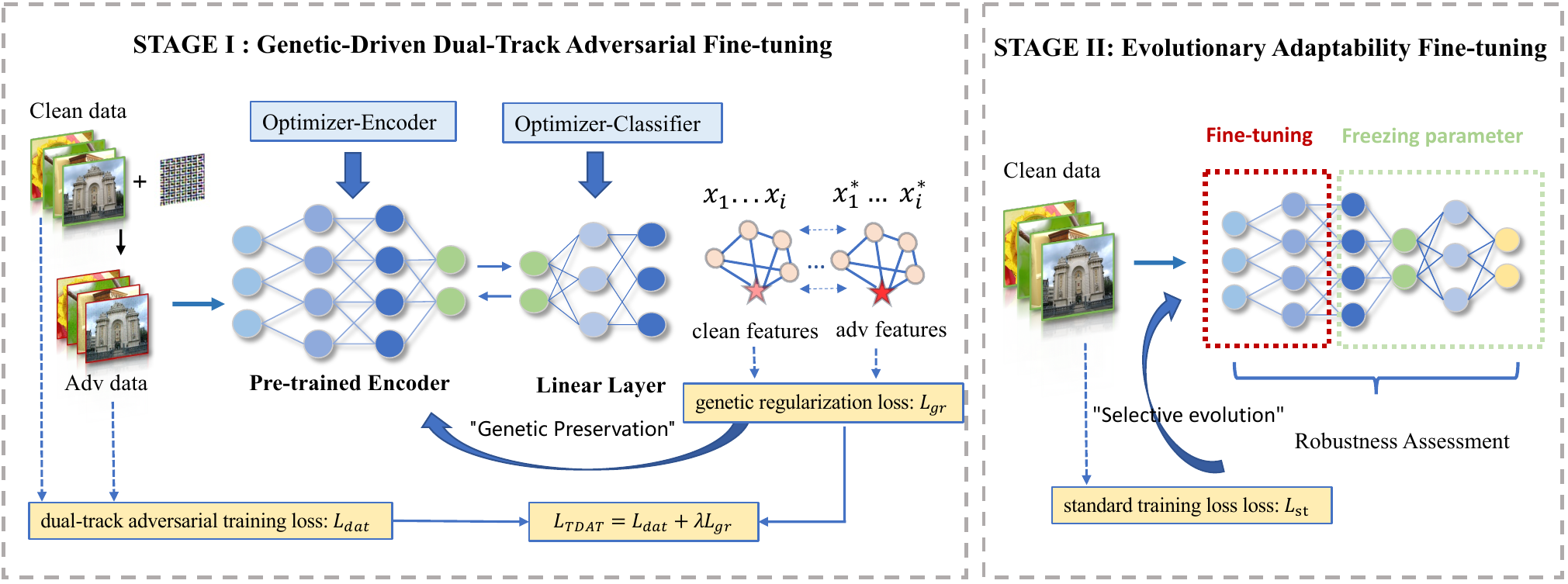}
    \caption{The pipeline of our defense}
    \label{fig:pipeline}
\end{figure*}

\section{Methodology}\label{sec:method}

\subsection{Problem Definition}
Let $ \mathcal{E}_{\theta_{e}} (\cdot )$ denote a well pre-trained encoder with parameters $\theta_{e}$. It takes an image $(x,y) \in \mathcal{D}_{p}$ as input and outputs a feature vector $v \in \mathcal{V}$, where $y$ represents the class label corresponding to $x$, $\mathcal{D}_{p}$ and $\mathcal{V}$ refer to the pre-training dataset and feature space, respectively.
By fine-tuning this encoder on the downstream dataset $\mathcal{D}_{d}$, a downstream model $\mathcal{M}_{\theta } (\cdot )$ (\ie, a pre-trained encoder $ \mathcal{E}_{\theta_{e}}$ combined with a classifier $\mathcal{F}_{\theta_{c} }$) can be constructed to accomplish the classification task, where $\theta$ and $\theta_{c}$ are parameters of downstream model and classifier.
The optimization goal of  downstream model fine-tuning can be expressed as:
\begin{equation} \label{eq:op1}
\mathop{\arg\min}\limits_{\theta_{e}, \theta_{c}}\mathbb{E}_{(x,y)\in \mathcal{D}}  \mathcal{L}_{CE} (\mathcal{F}(\theta_{c}, \mathcal{E}(\theta_{e}, x)), y)
\end{equation}
where $\mathcal{L}_{CE}$ denotes cross-entropy loss function.

The attacker employs a surrogate dataset $\mathcal{D}_a$ to generate a adversarial noise against the pre-trained encoder. 
The universal adversarial noise $\delta^{\ast} $ should be sufficiently small, and modeled through an upper-bound $\epsilon$ on the $l_{p}$-norm.
This problem can be formulated as:
\begin{equation} \label{eq:op2}
 \mathcal{E}_{\theta_{e}}\left (  x +  \delta^{\ast}   \right )  \ne  \mathcal{E}_{\theta_{e}}\left (  x  \right ), \quad  s.t.\left \| \delta^{\ast}   \right \| _{p}\le \epsilon
\end{equation}

The attacker's ultimate goal is to 
employ the noise $\delta$, attaching it to the downstream data $x \in \mathcal{D}_{d}$ to create adversarial examples, aimed at deceiving the downstream classifier $F$. 
This can be formalized as:
\begin{equation} \label{eq:op3}
\mathcal{F}_{\theta_{c}}   ( \mathcal{E}_{\theta_{e}}\left (  x + \delta^{\ast}   \right )) \ne \mathcal{F}_{\theta_{c}}(\mathcal{E}_{\theta_{e}}\left (  x  \right ) ), \quad s.t.  \left \| \delta^{\ast}   \right \|_{ p} \le \epsilon
\end{equation}

Defenders aim to enhance the robustness of downstream models through adversarial training, which involves adding perturbations $\delta \in \Omega$ to the training data, where $\Omega$ denotes  the allowed range of input perturbation.

\subsection{Gen-AF: A Complete Illustration}
In this section, we propose the first genetic evolution-eurtured adversarial fine-tuning (Gen-AF) to enhance the robustness of downstream models while preserving the inherent generalization capabilities of the pre-trained encoder.
The pipeline of Gen-AF is depicted in~\cref{fig:pipeline}, which consists of two stages: 
adversarial fine-tuning to enhance robustness and standard training to improve generalization.
The detailed optimization process is presented in Alg.~\ref{optimization_daf}.

\vpara{STAGE I. Genetic-Driven Dual-Track Adversarial Fine-tuning.}
The prerequisite for constructing the downstream model is to ensure the proper inheritance of parameters from the pre-trained model. Therefore, recognizing the significance of genetic information in the pre-trained encoder, in the first stage, we design the genetic-driven dual-track adversarial fine-tuning (GDAF) method to enhance robustness through adversarial fine-tuning while constraining the loss of model genes.
The loss function is as follows:
\begin{equation} \label{eq:m1}
\mathcal{L}_{GDAT} =  \mathcal{L}_{dat} +  \lambda  \mathcal{L}_{gr}
\end{equation}
where $\mathcal{L}_{dat}$ is the dual-track adversarial training loss and $\mathcal{L}_{gr}$ denotes the genetic regularization loss.

Based on \cref{sec:initution}, we design a genetic regularization loss that constrains the offset between adversarial examples and their corresponding benign samples in the feature space. This aims to prevent significant changes in feature boundaries that occur when using adversarial examples as training data, thereby avoiding the loss of inherited knowledge in the pre-trained encoder.
We first construct a graph $\mathsf{G}$  to delineate the topology relationship between adversarial and benign examples in the feature space, preserving genetic information in the pre-trained model by limiting dissimilarities, thereby efficiently adapting to new data domains.
In the graph, the nodes $\mathcal{V}$ represent the feature vector of samples $X$, the edges signify the relationships between the samples, and the edge weights $\mathcal{W}$  are determined by the degree of similarity between the feature vectors. 
The genetic regularization loss can be formalized as follows:
\begin{equation} \label{eq:m3}
\mathcal{L}_{gr} = \mathcal{L}_{CE} ( \mathsf{G}_{ben}(\mathcal{V}_{c}, \mathcal{S}_{c}; x),  \mathsf{G}_{adv}(\mathcal{V}_{a}, \mathcal{S}_{a}; x+\delta))
\end{equation}
where the $\mathsf{G}_{ben}$,  $\mathcal{V}_{c}$ and $\mathcal{S}_{c}$ represent the feature graph, node and edge set corresponding to the benign samples, respectively, while $\mathsf{G}_{adv}$, $\mathcal{V}_{a}$ and $\mathcal{S}_{a}$ denote the node and edge set corresponding to the adversarial examples.

To begin with, we input the benign and adversarial example (denoted as $x_{i}^{*}$) into the downstream model, yielding feature vector $v$ that assemble the graph’s nodes set $\mathcal{V}$, which can be depicted as:
\begin{equation} \label{eq:m4}
\mathcal{V} = \{ v_i \mid v_i = \mathcal{F}(\theta_{e}, \mathcal{E}(\theta_{e}, x_{i}^{*})), \forall i \in \{1, 2, \ldots, N\} \}  
\end{equation}

We then take the feature distances ($\mathcal{W}_{ij}$ stands for the distance between the node $v_{i}$ and $v_{j}$) between different nodes to represent the edges of the graph, with the edge set $\mathcal{S}$ being denoted as:
\begin{equation} \label{eq:m5}
\mathcal{S} = { (v_i, v_j, \mathcal{W}_{ij}), \quad 0 < i, j \le N, \quad i \neq j }
\end{equation}

To accurately measure the distance between two nodes (\eg, the $i_{th}$ and $j_{th}$ nodes), we consider employing cosine distance to quantify the dissimilarity in features between them. Additionally, we remove the nearest neighbor points to prevent the formation of isolated subgraphs caused by data points with excessively high local density, thereby ensuring the local connectivity of the manifold and better preserving the global structure.
We denote $\rho_{j}$ as the distance from the  $j_{th}$ note to its nearest neighbor.
This can be represented as:
\begin{equation} \label{eq:m6}
\mathcal{W}_{ij} = \frac{2 - (cos_{ij} - \rho_j)}{\sum_{k=1, k \ne j}^{N} (2 - (cos_{ik} - \rho_k))}
\end{equation}

Let $\mathcal{W}_{ij}^{c}$, $\mathcal{W}_{ij}^{a}$ represent  the feature distances between nodes in benign and adversarial samples, respectively. 
The genetic regularization loss can ultimately be formalized as:
\begin{equation} \label{eq:m7}
\mathcal{L}_{gr} =  \sum_{i}\sum_{j}\left [ \mathcal{W}_{ij}^{c} \log{(\frac{ \mathcal{W}_{ij}^{c}}{ \mathcal{W}_{ij}^{a}}) + (1-\mathcal{W}_{ij}^{c}) \log{(\frac{1- \mathcal{W}_{ij}^{c}}{1-\mathcal{W}_{ij}^{a}} })} \right ]
\end{equation}

We further propose a bilevel-optimizer collaborative strategy to optimize the parameters of the pre-trained encoder and the classifier separately, thereby enhancing the stability and efficiency of the adversarial fine-tuning process.
Specifically, during downstream adversarial fine-tuning, we first set a small learning rate $LR_{E}$ for the pre-trained encoder to ensure minimal adjustments to previously learned features, preventing the loss of the pre-trained encoder's inherent powerful feature extraction capabilities.
Simultaneously, a standard learning rate $LR_{C}$ is applied to the classifier, facilitating rapid model convergence.
We can formalize the the dual-track adversarial training loss $\mathcal{L}_{dat}$ as follows:
\begin{align} \label{eq:m2}
\begin{split}
\mathcal{L}_{dat}(x ,y;\theta_{e}, \theta_{c}) =  \min_{\delta \in \Omega}\mathcal{L}_{CE} (\mathcal{F}(\theta_{c}, \mathcal{E}(\theta_{e}, x+\delta)), y) 
\end{split}
\end{align}

\vpara{STAGE II. Evolutionary Adaptability Fine-tuning.}
In the second stage, we introduce an evolutionary adaptability fine-tuning (EAT) method, focusing on redundant genes to boost the initial model's generalization, which obtained in the first stage. 
Inspired by~\cite{chatterji2019intriguing, zhu2023improving},
we first calculate the sensitivity of each layer of the model to adversarial noise and create a sensitivity dictionary by ordering them according to their level of sensitivity.
Specifically, for an adversarially trained model, we modify the parameters of just one layer at a time and record the changes in the model's adversarial training loss values. This allows us to determine their robustness contribution (RC), which is then documented in a dictionary ($D_{rc}$).
This process can be represented as follows:

\begin{equation} \label{eq:m8}
\mathcal{L}_{drc} = \max \limits_{\Delta \theta \in \epsilon_{\theta}}\mathcal{L}_{dat}(x,y;\theta+\Delta \theta) - \mathcal{L}_{dat}(x,y;\theta)
\end{equation}
where 
$\theta$ represents the parameters of the entire model, including $\theta_{e}$ of the pre-trained encoder and $\theta_{e}$ of the classifier.
$\Delta \theta = \{\mathbf{0}, \hdots, \mathbf{0}, \Delta \theta^{(i)}, \mathbf{0}, \hdots, \mathbf{0}\}$ denotes the weight perturbation with respect to the module weights $\theta^{(k)}$, $\epsilon_{\theta} = \{ \Delta \theta \, \big|  \, \lVert \Delta \theta \rVert_p \leq \gamma \lVert \theta^{(k)} \rVert_p \}$.

After obtaining the model's layer-level robustness contribution dictionary $D_{rc}$ through \cref{eq:m8}, we select the top-k layers with the lowest robustness and acquire their parameters $\tilde{\Theta}_{\text{top-k}}$. We then perform evolutionary adaptability fine-tuning to further enhance the model's generalization ability by fine-tuning the redundant layers. 
It may include: the trainable parameters of the pre-trained encoder and classifier, \(\tilde{\theta}_{e}\) and \(\tilde{\theta_{f}}\). We then utilize benign data for the  standard fine-tuning.
The optimization objective is as follows:
\begin{equation} \label{eq:m10}
\mathcal{L}_{SAT} = \mathop{\arg\min}\limits_{\theta_{e}, \theta_{c}}\mathbb{E}_{(x,y)\in D}  \mathcal{L}_{CE} (x ,y;\tilde{\theta}_{e}, \tilde{\theta}_{c})
\end{equation}

\begin{algorithm}[H]
    \caption{Gen-AF}
    \label{optimization_daf}
    \begin{algorithmic}[1] 
        \REQUIRE training dataset  $ (x,y) \in \mathcal{D}_{d}$, pre-trained encoder $ \mathcal{E}$ with parameter $\theta_{e}$, downstream classifier $\mathcal{F}$ with parameter $\theta_{c}$, the number of training epochs $T$, loss weight $\lambda$, optimizer $Adam_{e}$, $Adam_{f}$, $Adam_{r}$
         max-perturbation constraint $\epsilon$. 
        \ENSURE roubst downstream model $ \mathcal{E}_{\theta_{e}}(\mathcal{F}_{\theta_{c}})$. 
        \STATE {Initialize adversarial dataset: $\mathcal{D}_{adv} = \{\}$.}
        \FOR{$x \in \mathcal{D}_{d}$}  
        \STATE {$x_{adv} = PGD(\theta_{e}, \theta_{c},  x)$} 
        \STATE {$\mathcal{D}_{adv} = \mathcal{D}_{adv} \bigcup x_{adv} $}
        \ENDFOR
        \STATE {\# STAGE I. Adversarial Fine-tuning} 
        \WHILE {max iterations or not  converge}
            \STATE {Calculate $\mathcal{L}_{GDAT}$ with Eq.~\ref{eq:m1}}
            \STATE {$\theta_{e} \gets Adam_{e}(\theta_{e}, \mathcal{L}_{GDAT}) $}
            \STATE {$\theta_{c} \gets Adam_{c}(\theta_{c}, \mathcal{L}_{GDAT}) $} 
            \STATE {Update $\mathcal{E}$ and $\mathcal{F}$ through backprop}
        \ENDWHILE
         \STATE {\# STAGE II. Standard Fine-tuning} 
         \STATE {Calculate RC $D_{rc}$ with Eq.~\ref{eq:m8}}
         \STATE {Select the parameters $\tilde{\Theta}_{\text{top-k}}$ of the top-k least robust layers from $D_{rc}$.}
         \STATE {Freeze all parameters of $\theta_{e}$ and $\theta_{c}$ except for $\tilde{\Theta}_{\text{top-k}}$}
         \WHILE {max iterations or not  converge}
            \STATE {Calculate $\mathcal{L}_{SAT}$ with Eq.~\ref{eq:m10}}
            \STATE {$\theta_{e}, \theta_{c}  \gets Adam_{r}(\theta_{e}, \theta_{c} ,\mathcal{L}_{SAT}) $}
            \STATE {Update $\mathcal{E}$ and $\mathcal{F}$ through backprop}
        \ENDWHILE
    \end{algorithmic} 
\end{algorithm}

\section{Experiments}\label{sec:experiment}

\begin{table*}[htbp]
  \centering
  \caption{The TA (\%) of  models based on standard training under different settings.}
    \scalebox{0.83}{
    \begin{tabular}{cccccccccccc}
    \toprule
     \rowcolor[gray]{0.9} \textbf{P-Dataset} & \textbf{Dataset} & \textbf{BYOL} &  \textbf{DINO} & \textbf{MoCo2+} & \textbf{MoCo3} & \textbf{NNCLR} & \textbf{RESSL} & \textbf{SimCLR} & \textbf{SwAV} & \textbf{VibCreg} & \textbf{W-MSR} \\
    \midrule
    \multirow{6}[2]{*}{CIFAR10} & ANIMALS10 & 83.22  & 80.21  & 81.90  & 82.37  & 81.32  & 79.93  & 77.18  & 78.40  & 81.55  & 73.60  \\
          & CIFAR10 & 94.46  & 91.41  & 94.97  & 94.76  & 93.84  & 92.74  & 93.35  & 92.50  & 92.26  & 90.21  \\
          & GTSRB & 93.87  & 95.76  & 93.63  & 93.94  & 96.18  & 94.14  & 94.96  & 97.64  & 96.82  & 91.68  \\
          & ImageNet20 & 60.62  & 60.04  & 59.68  & 59.54  & 58.19  & 58.24  & 54.64  & 54.34  & 55.85  & 51.47  \\
          & STL10 & 83.23  & 82.92  & 84.73  & 83.95  & 83.64  & 81.85  & 81.99  & 81.52  & 81.98  & 78.50  \\
          & SVHN  & 94.52  & 91.64  & 95.14  & 94.74  & 93.82  & 91.96  & 93.67  & 92.06  & 92.62  & 90.29  \\
    \midrule
    \multirow{6}[2]{*}{ImageNet} & ANIMALS10 & 65.01  & 65.00  & 63.92  & 64.83  & 63.10  & 64.08  & 59.96  & 60.54  & 63.37  & 59.93  \\
          & CIFAR10 & 72.21  & 72.92  & 72.25  & 72.41  & 73.46  & 72.86  & 69.23  & 71.04  & 72.67  & 68.53  \\
          & GTSRB & 88.01  & 91.27  & 87.55  & 86.73  & 91.45  & 88.22  & 87.68  & 90.15  & 91.36  & 88.19  \\
          & ImageNet20 & 75.14  & 75.21  & 73.42  & 72.73  & 74.52  & 71.73  & 67.67  & 71.20  & 71.48  & 66.15  \\
          & STL10 & 63.88  & 65.16  & 64.98  & 62.60  & 65.84  & 64.79  & 62.81  & 63.75  & 65.23  & 60.84  \\
          & SVHN  & 70.68  & 72.99  & 71.25  & 71.38  & 72.71  & 71.85  & 69.55  & 70.46  & 72.61  & 67.82  \\
    \bottomrule
    \end{tabular}%
    }
 \label{tab:clean_performance}%
\end{table*}%

\begin{table*}[htbp]
  \centering
  \caption{The RA (\%) of adversarially trained models with Gen-AF under partial upstream-knowledge attacker settings.  $\mathcal{D}_{1}$ - $\mathcal{D}_{6}$ denote the settings where the downstream datasets are ANIMALS10, CIFAR10, GTSRB,  ImageNet20, STL10, and SVHN, respectively.}
\tabcolsep=0.3cm
  \scalebox{0.7}{
    \begin{tabular}{cccccccccccccc}
    \toprule[1.5pt]
 \rowcolor[gray]{0.9} \textbf{P-Dataset} & \textbf{Method}  & \textbf{Dataset} & \textbf{BYOL} & \textbf{DINO} & \textbf{MoCo2+} & \textbf{MoCo3} & \textbf{NNCLR} & \textbf{RESSL} & \textbf{SimCLR} & \textbf{SwAV} & \textbf{VibCreg} & \textbf{W-MSR} \\
    \midrule
    \multirow{35}[10]{*}{\rotatebox{90}{\large CIFAR10}}  & \multirow{7}[2]{*}{UAP~\cite{moosavi2017universal}} & $\mathcal{D}_{1}$ & 91.80 & 70.76 & \textbf{95.07} & 94.07 & 87.70 & 93.87 & 90.93 & 90.27 & 89.06 & 81.13 \\
          &       & $\mathcal{D}_{2}$ & 87.22 & 76.37 & \textbf{87.80} & 87.72 & 78.45 & 81.95 & 86.03 & 80.12 & 76.03 & 71.10 \\
          &       & $\mathcal{D}_{3}$ & \textbf{99.90} & 99.73 & 98.09 & 98.94 & 99.86 & 99.86 & 99.78 & 99.83 & 99.84 & 98.85 \\
          &       & $\mathcal{D}_{4}$ & 84.17 & 67.73 & \textbf{86.18} & 84.52 & 73.47 & 83.05 & 82.27 & 56.91 & 76.37 & 71.68 \\
          &       & $\mathcal{D}_{5}$ & 60.67 & 67.09 & 68.99 & 78.31 & \textbf{80.05} & 76.55 & 78.65 & 72.20 & 77.91 & 75.31 \\
          &       & $\mathcal{D}_{6}$ & 85.64 & 81.92 & \textbf{87.60} & 86.39 & 87.43 & 84.43 & 85.28 & 85.75 & 86.37 & 76.74 \\
          &       & AVG   & 84.90 & 77.27 & 87.29 & \textbf{88.32} & 84.49 & 86.62 & 87.16 & 80.85 & 84.26 & 79.13 \\
\cmidrule{2-13}          & \multirow{7}[2]{*}{UAPGD~\cite{deng2020universal}} & $\mathcal{D}_{1}$ & 91.89 & 70.79 & \textbf{95.31} & 94.82 & 88.08 & 93.86 & 90.73 & 90.90 & 88.75 & 85.77 \\
          &       & $\mathcal{D}_{2}$ & 86.90 & 73.08 & \textbf{88.24} & 87.71 & 78.47 & 82.12 & 85.91 & 80.53 & 75.76 & 74.48 \\
          &       & $\mathcal{D}_{3}$ & 99.72 & 99.54 & 98.60 & 99.75 & 99.73 & 99.76 & 99.76 & 99.82 & \textbf{99.83} & 98.78 \\
          &       & $\mathcal{D}_{4}$ & 82.89 & 66.24 & 86.41 & \textbf{86.48} & 74.87 & 82.77 & 82.89 & 67.40 & 75.66 & 74.11 \\
          &       & $\mathcal{D}_{5}$ & 60.46 & 66.72 & 68.71 & \textbf{81.93} & 80.19 & 76.99 & 80.27 & 77.61 & 77.33 & 79.04 \\
          &       & $\mathcal{D}_{6}$ & 85.46 & 81.96 & \textbf{87.89} & 86.84 & 87.46 & 84.28 & 86.05 & 85.90 & 85.47 & 78.77 \\
          &       & AVG   & 84.55 & 76.39 & 87.53 & \textbf{89.59} & 84.80 & 86.63 & 87.60 & 83.69 & 83.80 & 81.83 \\
\cmidrule{2-13}          & \multirow{7}[2]{*}{SSP~\cite{naseer2020self}} & $\mathcal{D}_{1}$ & 92.11 & 41.85 & \textbf{94.93} & 94.32 & 50.78 & 94.09 & 90.49 & 86.15 & 84.31 & 70.12 \\
          &       & $\mathcal{D}_{2}$ & 87.34 & 62.44 & \textbf{87.48} & 87.47 & 78.55 & 82.07 & 85.97 & 62.29 & 75.42 & 26.73 \\
          &       & $\mathcal{D}_{3}$ & 99.88 & 99.17 & 98.36 & 98.89 & 99.13 & \textbf{99.89} & 99.87 & 99.70 & \textbf{99.89} & 96.68 \\
          &       & $\mathcal{D}_{4}$ & 65.46 & 58.23 & 84.90 & \textbf{85.34} & 16.42 & 79.11 & 82.09 & 21.53 & 71.78 & 60.79 \\
          &       & $\mathcal{D}_{5}$ & 60.66 & 39.18 & 68.62 & 74.11 & \textbf{74.43} & 70.58 & 68.93 & 37.28 & 43.30 & 67.57 \\
          &       & $\mathcal{D}_{6}$ & 85.67 & 81.10 & 87.59 & 86.93 & \textbf{87.70} & 84.53 & 86.12 & 86.06 & 81.90 & 31.61 \\
          &       & AVG   & 81.85 & 63.66 & 86.98 & \textbf{87.84} & 67.83 & 85.04 & 85.58 & 65.50 & 76.10 & 58.92 \\
\cmidrule{2-13}          & \multirow{7}[2]{*}{PAP~\cite{ban2022pre}} & $\mathcal{D}_{1}$ & 92.12 & 34.39 & \textbf{95.06} & 93.54 & 44.36 & 94.07 & 90.57 & 82.48 & 50.40 & 55.42 \\
          &       & $\mathcal{D}_{2}$ & 87.65 & 75.05 & \textbf{89.30} & 86.98 & 78.82 & 82.16 & 86.11 & 58.33 & 72.02 & 54.70 \\
          &       & $\mathcal{D}_{3}$ & \textbf{99.93} & 99.73 & 99.54 & 97.45 & 99.76 & 99.88 & 99.88 & 99.53 & 99.90 & 96.86 \\
          &       & $\mathcal{D}_{4}$ & 36.41 & 46.37 & \textbf{86.14} & 82.64 & 14.20 & 78.66 & 81.98 & 42.59 & 58.86 & 55.92 \\
          &       & $\mathcal{D}_{5}$ & 60.45 & 31.34 & 68.69 & \textbf{73.23} & 71.97 & 72.68 & 52.01 & 58.46 & 44.93 & 63.02 \\
          &       & $\mathcal{D}_{6}$ & 85.64 & 81.44 & 87.79 & 85.38 & \textbf{88.08} & 84.51 & 85.89 & 83.96 & 80.65 & 60.32 \\
          &       & AVG   & 77.03 & 61.39 & \textbf{87.75} & 86.54 & 66.20 & 85.33 & 82.74 & 70.89 & 67.80 & 64.37 \\
\cmidrule{2-13}          & \multirow{7}[2]{*}{AdvEncoder~\cite{zhou2023advencoder}} & $\mathcal{D}_{1}$ & 91.48 & 65.68 & \textbf{94.45} & 94.09 & 83.14 & 91.43 & 88.13 & 89.91 & 86.09 & 71.48 \\
          &       & $\mathcal{D}_{2}$ & 86.53 & 34.71 & 64.15 & \textbf{87.55} & 75.62 & 78.09 & 82.93 & 78.88 & 73.79 & 78.09 \\
          &       & $\mathcal{D}_{3}$ & 68.12 & 98.79 & 94.59 & 99.23 & 96.96 & 98.81 & 95.47 & \textbf{99.71} & 91.92 & 98.48 \\
          &       & $\mathcal{D}_{4}$ & 82.00 & 57.16 & 83.60 & \textbf{86.39} & 72.12 & 78.43 & 78.25 & 49.88 & 71.44 & 69.26 \\
          &       & $\mathcal{D}_{5}$ & 60.47 & 58.13 & 66.61 & 70.47 & 73.40 & 63.57 & \textbf{77.49} & 63.45 & 69.67 & 70.89 \\
          &       & $\mathcal{D}_{6}$ & 85.37 & 77.64 & 81.30 & \textbf{86.57} & 67.99 & 72.51 & 81.58 & 85.68 & 49.44 & 77.90 \\
          &       & AVG   & 79.00 & 65.35 & 80.78 & \textbf{87.38} & 78.20 & 80.47 & 83.90 & 77.92 & 73.73 & 77.68 \\
    \bottomrule[1.5pt]
    \end{tabular}%
    }
  \label{tab:ra_our_cifar10}%
\end{table*}%

\begin{table*}[t]
  \centering
  \caption{The RA (\%) of adversarially trained models with Gen-AF under partial upstream-knowledge attacker settings.}
  \tabcolsep=0.3cm
  \scalebox{0.7}{
    \begin{tabular}{ccccccccccccccc}
    \toprule[1.5pt]
 \rowcolor[gray]{0.9} \textbf{P-Dataset} & \textbf{Method}  & \textbf{Dataset} & \textbf{BYOL} & \textbf{DINO} & \textbf{MoCo2+} & \textbf{MoCo3} & \textbf{NNCLR} & \textbf{RESSL} & \textbf{SimCLR} & \textbf{SwAV} & \textbf{VibCreg} & \textbf{W-MSR} \\
    \midrule
    \multirow{35}[10]{*}{\rotatebox{90}{\large ImageNet}}   & \multirow{7}[2]{*}{UAP~\cite{moosavi2017universal}} & $\mathcal{D}_{1}$ & \textbf{91.11} & 73.75 & 90.38 & 89.57 & 90.72 & 81.94 & 89.01 & 68.36 & 79.34 & 49.28 \\
          &       & $\mathcal{D}_{2}$ & 63.92 & 67.19 & 46.78 & 69.11 & 67.57 & 71.61 & 66.85 & \textbf{73.98} & 64.70 & 68.87 \\
          &       & $\mathcal{D}_{3}$ & 99.39 & 98.88 & 99.25 & 97.20 & 97.29 & 99.24 & \textbf{99.40} & 98.76 & 98.86 & 99.28 \\
          &       & $\mathcal{D}_{4}$ & 76.67 & 75.37 & \textbf{85.62} & 84.83 & 78.98 & 78.73 & 71.94 & 61.89 & 81.13 & 56.81 \\
          &       & $\mathcal{D}_{5}$ & \textbf{68.30} & 53.78 & 64.98 & 60.30 & 61.99 & 59.83 & 64.43 & 45.57 & 61.00 & 58.73 \\
          &       & $\mathcal{D}_{6}$ & 66.87 & 64.52 & 66.50 & 72.20 & \textbf{74.18} & 70.11 & 63.34 & 69.49 & 73.40 & 54.05 \\
          &       & AVG   & 77.71 & 72.25 & 75.59 & \textbf{78.87} & 78.45 & 76.91 & 75.83 & 69.68 & 76.40 & 64.50 \\
\cmidrule{2-13}          & \multirow{7}[2]{*}{UAPGD~\cite{deng2020universal}} & $\mathcal{D}_{1}$ & 91.97 & 84.54 & 90.75 & 92.65 & \textbf{92.73} & 82.75 & 90.44 & 79.44 & 90.21 & 79.54 \\
          &       & $\mathcal{D}_{2}$ & 70.51 & 78.13 & 62.87 & 70.41 & 68.77 & \textbf{79.50} & 66.58 & 78.83 & 77.06 & 74.99 \\
          &       & $\mathcal{D}_{3}$ & 99.45 & 99.21 & 99.49 & 99.32 & 99.69 & 99.41 & \textbf{99.77} & 99.41 & 99.35 & 99.37 \\
          &       & $\mathcal{D}_{4}$ & 77.33 & 81.64 & 86.28 & \textbf{86.44} & 81.59 & 79.09 & 72.48 & 68.92 & 86.34 & 65.94 \\
          &       & $\mathcal{D}_{5}$ & 69.59 & 64.98 & 68.15 & 62.79 & 66.98 & 67.81 & 68.65 & \textbf{71.88} & 40.12 & 68.44 \\
          &       & $\mathcal{D}_{6}$ & 70.14 & 76.48 & 66.85 & 78.90 & \textbf{80.98} & 77.91 & 63.81 & 76.44 & 79.13 & 67.38 \\
          &       & AVG   & 79.83 & 80.83 & 79.06 & 81.75 & \textbf{81.79} & 81.08 & 76.96 & 79.15 & 78.70 & 75.94 \\
\cmidrule{2-13}          & \multirow{7}[2]{*}{SSP~\cite{naseer2020self}} & $\mathcal{D}_{1}$ & \textbf{90.87} & 73.33 & 89.94 & 87.75 & 87.70 & 82.68 & 90.02 & 60.61 & 79.96 & 57.79 \\
          &       & $\mathcal{D}_{2}$ & 51.19 & 54.44 & 60.20 & 68.89 & 67.11 & 76.82 & 66.74 & \textbf{79.25} & 62.17 & 60.06 \\
          &       & $\mathcal{D}_{3}$ & 99.27 & 97.78 & 98.91 & 93.94 & 93.39 & 99.38 & \textbf{99.67} & 99.11 & 96.81 & 98.52 \\
          &       & $\mathcal{D}_{4}$ & 76.65 & 74.55 & \textbf{85.27} & 83.69 & 77.65 & 78.98 & 72.42 & 57.64 & 80.45 & 56.86 \\
          &       & $\mathcal{D}_{5}$ & \textbf{66.82} & 56.40 & 59.41 & 59.39 & 60.23 & 49.63 & 65.92 & 35.18 & 57.09 & 61.13 \\
          &       & $\mathcal{D}_{6}$ & 56.75 & 43.77 & 66.46 & 69.35 & 68.70 & 71.72 & 63.23 & \textbf{79.16} & 73.51 & 49.64 \\
          &       & AVG   & 73.59 & 66.71 & 76.70 & \textbf{77.17} & 75.80 & 76.54 & 76.33 & 68.49 & 75.00 & 64.00 \\
\cmidrule{2-13}          & \multirow{7}[2]{*}{PAP~\cite{ban2022pre}} & $\mathcal{D}_{1}$ & \textbf{90.26} & 75.59 & 89.07 & 87.87 & 89.95 & 81.60 & 88.32 & 72.36 & 82.18 & 30.41 \\
          &       & $\mathcal{D}_{2}$ & 56.72 & 74.02 & 63.29 & 67.64 & 67.64 & \textbf{76.92} & 65.53 & 70.16 & 69.27 & 69.17 \\
          &       & $\mathcal{D}_{3}$ & 91.33 & 92.85 & 95.61 & 86.96 & 92.60 & 92.73 & 89.62 & 83.22 & 76.57 & \textbf{98.95} \\
          &       & $\mathcal{D}_{4}$ & 74.08 & 76.31 & \textbf{84.24} & 80.97 & 78.16 & 78.34 & 70.11 & 64.11 & 79.17 & 52.88 \\
          &       & $\mathcal{D}_{5}$ & \textbf{67.46} & 61.52 & 60.47 & 60.49 & 64.36 & 50.13 & 65.48 & 50.79 & 65.81 & 57.03 \\
          &       & $\mathcal{D}_{6}$ & 62.44 & 72.76 & 66.61 & \textbf{75.88} & 73.46 & 72.46 & 62.62 & 65.61 & 74.98 & 48.34 \\
          &       & AVG   & 73.71 & 75.51 & 76.55 & 76.63 & \textbf{77.69} & 75.36 & 73.61 & 67.71 & 74.66 & 59.46 \\
\cmidrule{2-13}          & \multirow{7}[2]{*}{AdvEncoder~\cite{zhou2023advencoder}} & $\mathcal{D}_{1}$ & 88.58 & 67.25 & \textbf{90.01} & 84.23 & 87.92 & 80.31 & 84.41 & 64.69 & 78.89 & 56.57 \\
          &       & $\mathcal{D}_{2}$ & 59.40 & 58.25 & 53.66 & 67.04 & 65.41 & 67.20 & 65.29 & 58.51 & 63.28 & \textbf{70.63} \\
          &       & $\mathcal{D}_{3}$ & \textbf{99.39} & 98.96 & 99.30 & 95.25 & 83.77 & 99.09 & 82.71 & 97.53 & 97.48 & 99.35 \\
          &       & $\mathcal{D}_{4}$ & 74.36 & 66.40 & \textbf{84.04} & 78.48 & 73.24 & 77.35 & 69.17 & 57.55 & 79.92 & 50.39 \\
          &       & $\mathcal{D}_{5}$ & 65.64 & 43.73 & \textbf{66.00} & 53.66 & 56.09 & 59.85 & 62.28 & 47.46 & 57.97 & 56.07 \\
          &       & $\mathcal{D}_{6}$ & 64.01 & 54.31 & 66.39 & 63.65 & 61.21 & 60.71 & 62.21 & 51.42 & \textbf{72.32} & 43.65 \\
          &       & AVG   & 75.23 & 64.82 & \textbf{76.57} & 73.72 & 71.27 & 74.08 & 71.01 & 62.86 & 74.98 & 62.78 \\
    \bottomrule[1.5pt]
    \end{tabular}%
    }
  \label{tab:ra_our_imagenet}%
\end{table*}%

\begin{table*}[htbp]
  \centering
 \caption{The TA (\%) of adversarially trained models with Gen-AF under different settings.}
    \scalebox{0.83}{
    \begin{tabular}{cccccccccccccc}
    \toprule
         \rowcolor[gray]{0.9} \textbf{P-Dataset} & \textbf{Dataset} & \textbf{BYOL} & \textbf{DINO} & \textbf{MoCo2+} & \textbf{MoCo3} & \textbf{NNCLR} & \textbf{RESSL} & \textbf{SimCLR} & \textbf{SwAV} & \textbf{VibCreg} & \textbf{W-MSR} \\
       \midrule
    \multirow{6}[2]{*}{CIFAR10} & ANIMALS10 & 91.96 & 86.05 & 95.32 & 95.22 & 91.69 & 94.07 & 91.11 & 92.17 & 89.23 & 87.42 \\
          & CIFAR10 & 87.99 & 88.57 & 90.14 & 88.02 & 78.55 & 82.02 & 86.09 & 88.97 & 75.96 & 88.48 \\
          & GTSRB & 99.98 & 99.85 & 99.97 & 99.91 & 99.87 & 99.90 & 99.92 & 99.88 & 99.92 & 99.92 \\
          & ImageNet20 & 85.00 & 74.47 & 86.21 & 87.56 & 84.40 & 84.42 & 83.30 & 75.44 & 77.29 & 76.23 \\
          & STL10 & 60.62 & 82.04 & 69.06 & 85.56 & 80.66 & 78.66 & 82.14 & 82.43 & 84.58 & 82.37 \\
          & SVHN  & 85.69 & 82.04 & 88.18 & 86.95 & 88.52 & 84.62 & 86.51 & 86.54 & 88.13 & 86.47 \\
    \midrule
    \multirow{6}[2]{*}{ImageNet} & ANIMALS10 & 92.09 & 86.71 & 90.86 & 92.89 & 93.20 & 83.13 & 90.79 & 84.49 & 92.02 & 87.89 \\
          & CIFAR10 & 71.72 & 81.12 & 65.03 & 70.69 & 68.96 & 81.89 & 66.89 & 81.27 & 79.51 & 76.29 \\
          & GTSRB & 99.46 & 99.33 & 99.60 & 99.79 & 99.86 & 99.52 & 99.84 & 99.59 & 99.65 & 99.49 \\
          & ImageNet20 & 77.79 & 81.91 & 86.33 & 86.51 & 82.11 & 79.80 & 72.33 & 70.77 & 86.82 & 67.54 \\
          & STL10 & 70.25 & 66.46 & 69.09 & 63.66 & 70.08 & 67.64 & 68.69 & 54.98 & 73.03 & 68.34 \\
          & SVHN  & 70.86 & 80.62 & 66.96 & 81.36 & 82.55 & 82.20 & 64.24 & 81.71 & 79.76 & 76.48 \\
    \bottomrule
    \end{tabular}%
    }
  \label{tab:ta_our}%
\end{table*}%

\subsection{Related Attacks and Defenses}
\textbf{Attacks.} 
We employ the following five universal adversarial attacks to assess the efficacy of our work.

\begin{itemize}

\item {\bf UAP~\cite{moosavi2017universal}:}
UAP introduces image-agnostic perturbations that deceive DNNs across various inputs with a single perturbation.

\item {\bf UAPGD~\cite{deng2020universal}:}
UAPGD enhances attack performance by integrating the PGD~\cite{carlini2017towards} algorithm into UAP.

\item {\bf SSP~\cite{naseer2020self}:}
SSP does not rely on decision boundary information and creates universal adversarial examples by directly perturbing the feature space through maximizing the feature loss of DNNs.

\item {\bf PAP~\cite{ban2022pre}:}
PAP designs pre-trained adversarial perturbations by lifting the feature activations of low-level layers. It includes three variant attacks: $L4_{base}$, $L4_{fuse}$, and $L4_{ugs}$, with $L4_{ugs}$ being the most potent.
We refer to $L4_{ugs}$  by default when using the term PAP.

\item {\bf AdvEncoder~\cite{zhou2023advencoder}:} 
AdvEncoder creates adversarial examples by altering the high-frequency components of the image, independent of any pre-trained dataset and downstream task information. It includes two forms of attacks: Adv-PER based on perturbations and Adv-PAT based on patches. We refer to Adv-PER by default when using the term AdvEncoder.

\end{itemize}

\textbf{Defenses.} 
We employ three state-of-the-art adversarial training mechanisms to assess the excellence of our work.
\begin{itemize}

\item {\bf PGD-AT~\cite{carlini2017towards}:}
PGD-AT proposes to use the PGD algorithm to generate adversarial examples for improving the robustness of the model.

\item {\bf TRADES~\cite{zhang2019theoretically}:}
TRADES focuses on balancing the trade-off between model accuracy on clean data and robustness against adversarial examples. This is achieved by minimizing a specially designed loss function that incorporates both the standard classification loss on clean examples and an additional term that penalizes the model when the output distribution significantly differs between clean and adversarial inputs. 

\item {\bf MART~\cite{wang2019improving}:}
MART introduces the misclassification aware adversarial training, which explicitly differentiates between misclassified and correctly classified samples during the training process. By adopting distinct strategies for handling these two types of samples, the approach significantly enhances the model's robustness against adversarial examples.
\end{itemize}

\subsection{Experimental Setting}\label{sec:experiment_setting}
\noindent\textbf{Datasets and Models.} 
We measure the performance of Gen-AF across the following six image benchmarks:
CIFAR10~\cite{krizhevsky2009learning}, STL10~\cite{coates2011analysis}, GTSRB~\cite{stallkamp2012man}, ImageNet20~\cite{russakovsky2015ImageNet}, SVHN~\cite{netzer2011reading},  ANIMALS10~\cite{Animals10}.
 We use the publicly available pre-trained encoders from \emph{solo-learn}\footnote{https://github.com/vturrisi/solo-learn}, an established SSL library as victim encoders. 
Following~\cite{zhou2023advencoder}, we select ten SSL methods (BYOL~\cite{grill2020bootstrap}, DINO~\cite{caron2021emerging},  MoCo v2+~\cite{chen2020improved}, MoCo v3~\cite{chen2021empirical}, NNCLR~\cite{dwibedi2021little}, ReSSL~\cite{zheng2021ressl}, SimCLR~\cite{chen2020simple},  SwAV~\cite{caron2020unsupervised}, VIbCReg~\cite{lee2021vibcreg},  W-MSE~\cite{ermolov2021whitening}). 
All the encoders are pre-trained on ImageNet~\cite{russakovsky2015ImageNet} or CIFAR10~\cite{krizhevsky2009learning} with ResNet18 as backbones.

\begin{figure*}[!t]   
  \centering
 \subcaptionbox{CIFAR10-ImageNet \label{a}}{\includegraphics[width=0.24\textwidth]{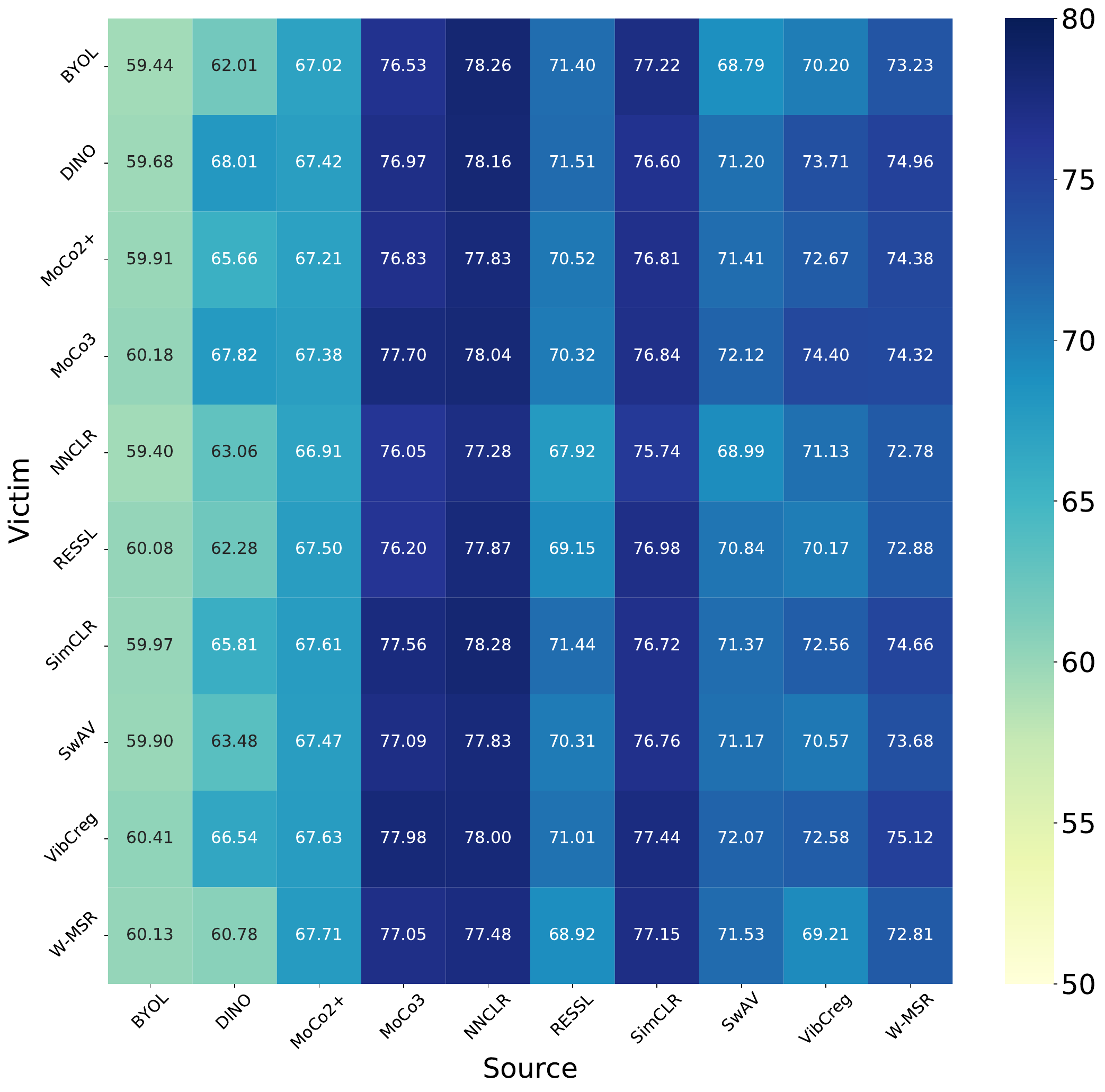}}
     \subcaptionbox{CIFAR10-CIFAR10\label{b}}{\includegraphics[width=0.24\textwidth]{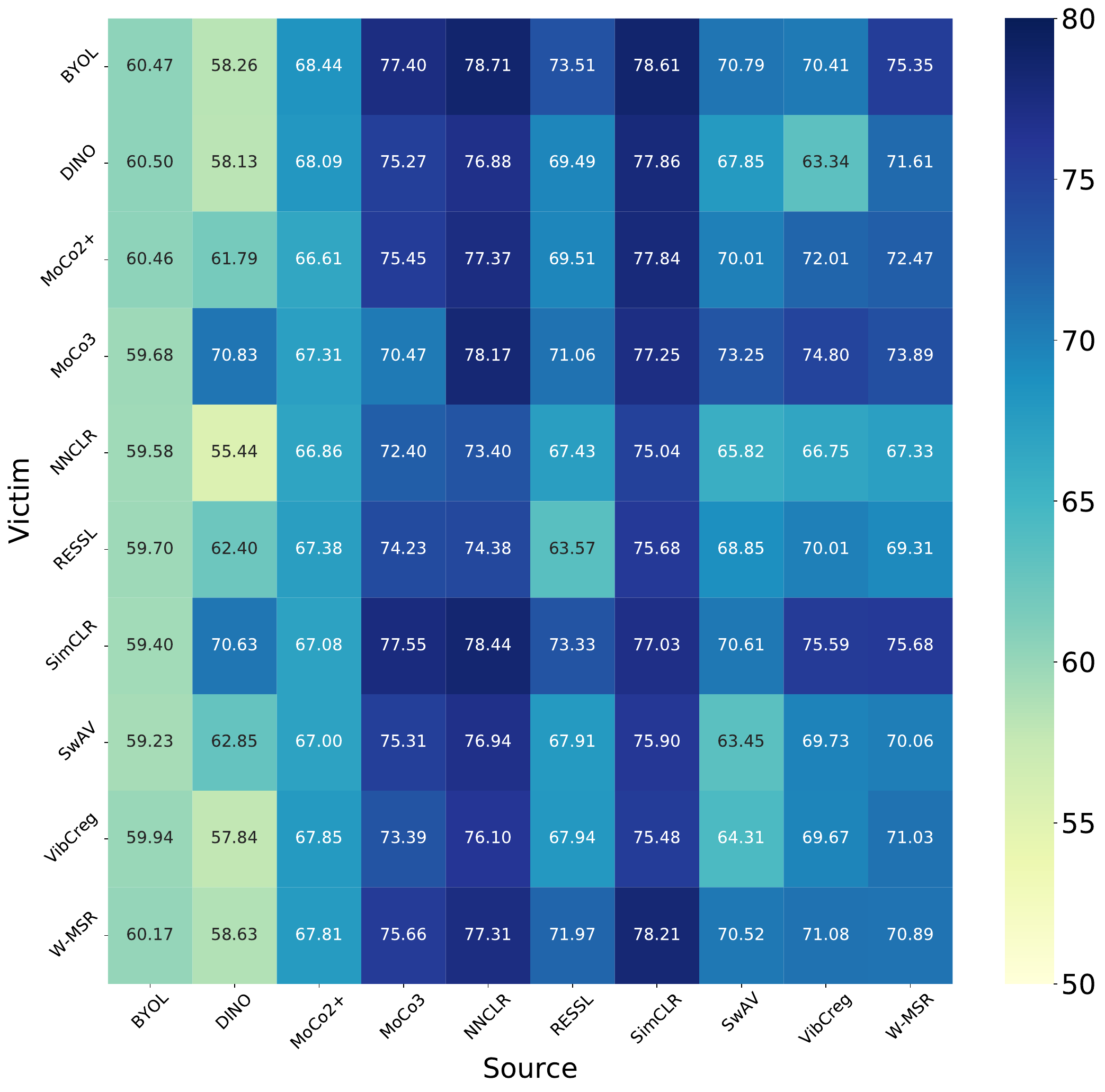}}
    \subcaptionbox{ImageNet-CIFAR10 \label{c}}{\includegraphics[width=0.244\textwidth]{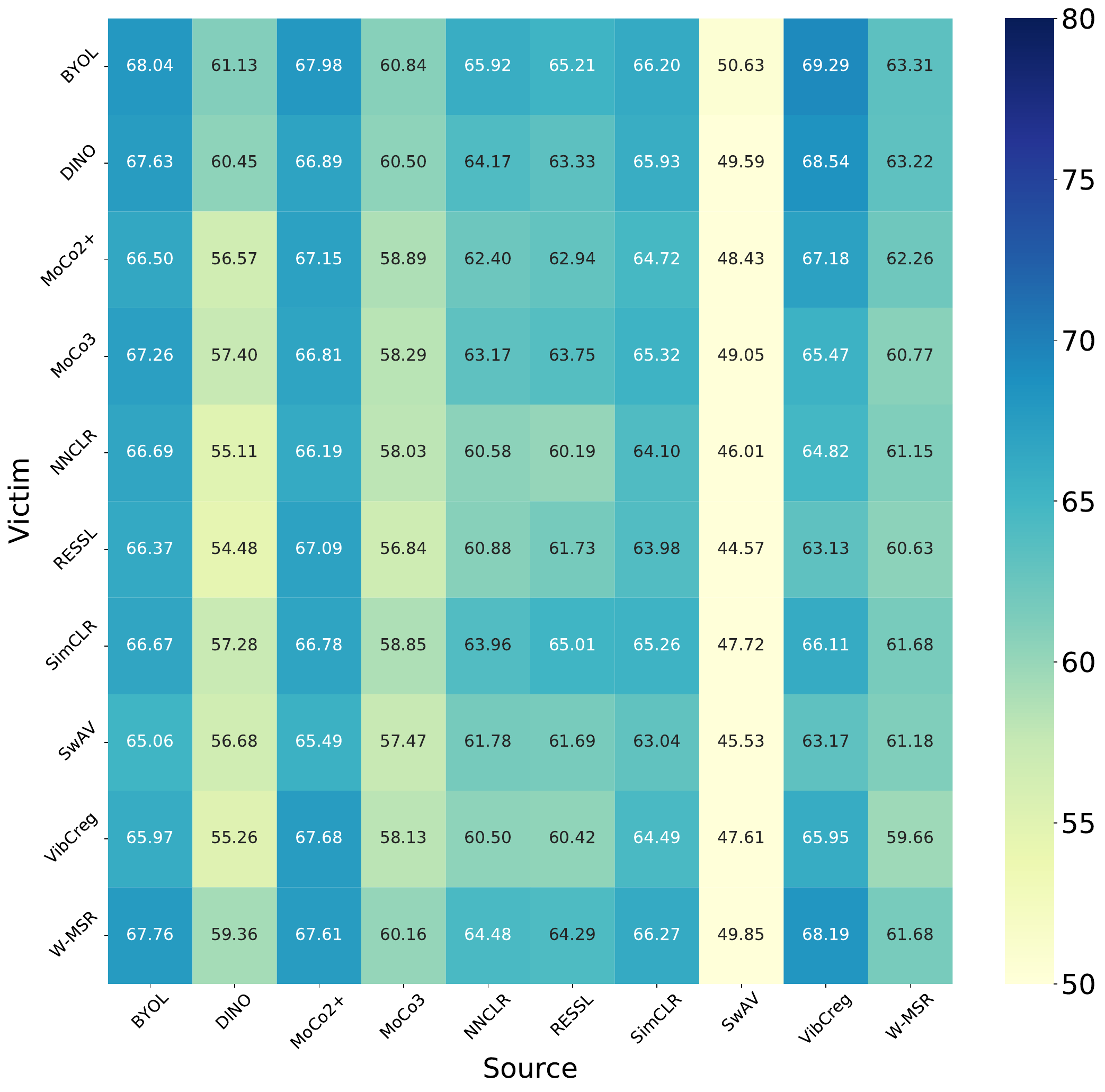}}
     \subcaptionbox{ImageNet-ImageNet \label{d}}{\includegraphics[width=0.243\textwidth]{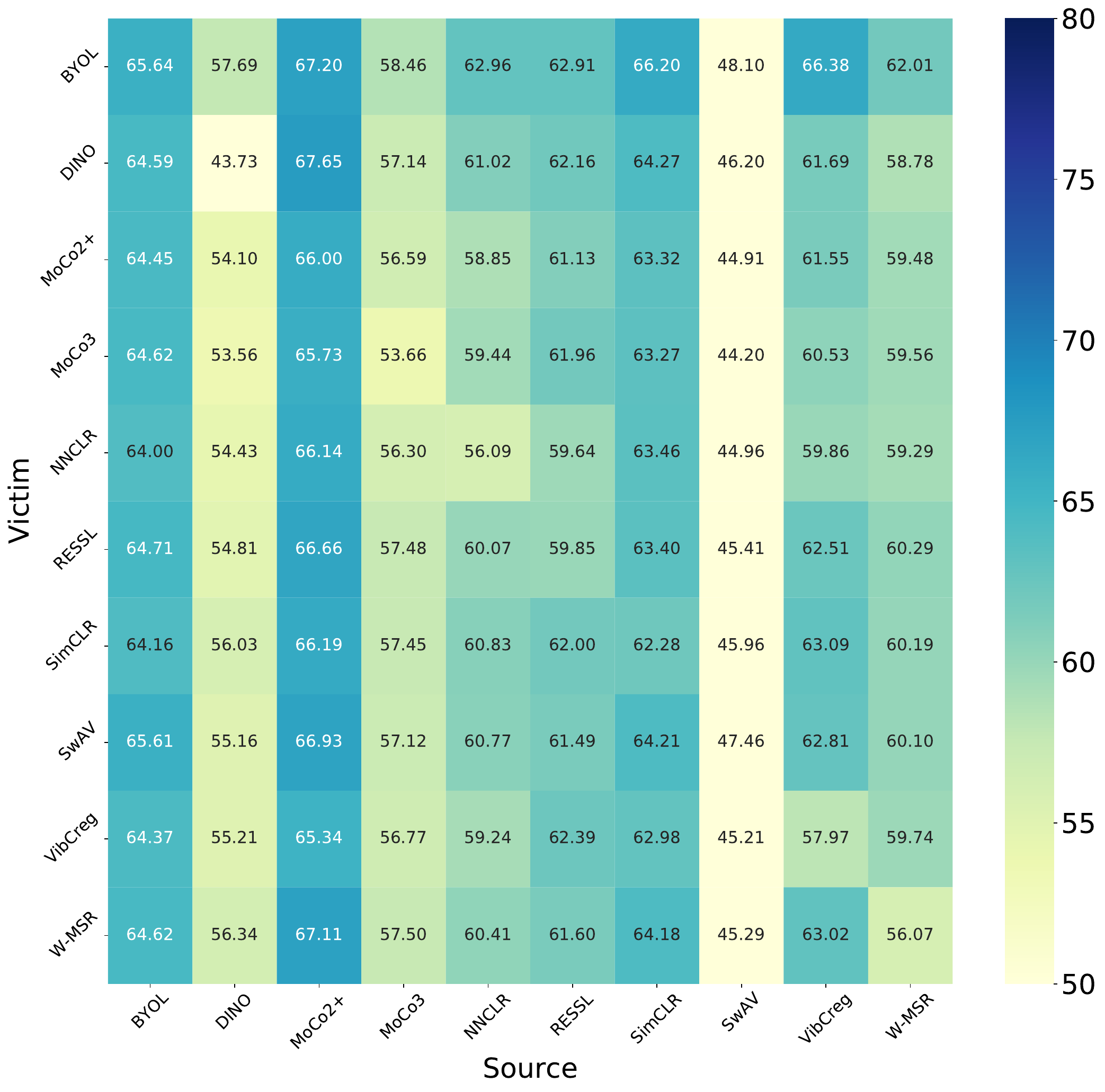}}
\caption{
The RA (\%) of adversarially trained downstream models with Gen-AF under transfer-based black-box attacker settings.
CIFAR10-ImageNet represents that we use CIFAR10 and ImageNet to train two encoders based on which adversarial examples and downstream tasks are made, respectively. Others have the same definition. (a) - (b) denote CIFAR10 pre-training results and (c) - (d) represent ImageNet pre-training results. }
\label{fig:black_box_transfer}
\end{figure*}

\noindent\textbf{Evaluation Metrics.} 
To assess the effectiveness of our proposed method, we use the following three metrics to evaluate the robustness and generalization of models after adversarial training.

\begin{itemize}

\item {\bf Testing Accuracy:} \textit{Standard Testing Accuracy} (TA) denotes the classification accuracy on the clean test dataset.  A higher value indicates stronger generalization capability.

\item {\bf Robust Testing Accuracy:} \textit{Robust Testing Accuracy} (RA) stands the classification accuracy on the adversarial examples.  A higher value indicates stronger robustness capability.

\item {\bf Attack Success Rate:} 
\textit{Attack Success Rate} (ASR) is a metric from the attacker's perspective, used to enhance our understanding of a model's robustness. It represents the rate of adversarial examples whose model predictions differ from their corresponding clean samples. A higher value indicates stronger attack capability.
\end{itemize}

\noindent\textbf{Implementation Details.} 
Following~\cite{zhou2023advencoder, ban2022pre}, we set the upper limit of perturbation $\epsilon$ to $10/255$ for perturbation-based attacks, and the patch size to $0.03$ for for patch-based attacks.
We select CIFAR10 as the default surrogate dataset for the attacker.
We set the hyper-parameters $\lambda = 20$ and the training epoch to $50$ with batch size of $256$. 
We employ the Adam optimizer for adversarial training. 
In the first stage, the learning rate for the pre-trained encoder is set to $0.0001$, and for the classifier, it is set to  $0.005$. In the second stage, we select the top $20\%$ of robust insensitive layers for fine-tuning and set the overall learning rate to $0.001$.
Our codes are available at: \url{https://github.com/CGCL-codes/Gen-AF}.

\subsection{Quasi-black-box Scenario}\label{sec:qbox}
In a quasi-black-box scenario,  we aim to defend against the full upstream-knowledge attacker and partial upstream-knowledge attacker. The key distinction between them is whether the attacker's substitute data is consistent with the pre-training dataset.
We assess our defense scheme using ten self-supervised pre-trained encoders on CIFAR10 or ImageNet across six downstream datasets, employing five UAP methods: AdvEncoder, PAP, UAP, UAPGD, and SSP, to evaluate our approach's effectiveness.
Among them, AdvEncoder is the SOTA downstream-agnostic adversarial examples method, allowing attackers to effectively fool downstream models without knowledge of the pre-training dataset or downstream tasks.

We provide the clean performance of downstream models based on these encoders in \cref{tab:clean_performance}.
When the attacker's surrogate dataset is consistent with the pre-training dataset (\ie, full upstream-knowledge attacker), the PAP exhibits the best overall performance. Conversely, when the surrogate dataset does not align with the pre-training dataset (\ie,  partial upstream-knowledge attacker), the AdvEncoder outperforms the others. This aligns with the underlying assumptions emphasized by these two types of attacks.

As shown in \cref{tab:ra_our_cifar10} and \cref{tab:ra_our_imagenet}, we evaluate the performance of our proposed method in quasi-black-box scenarios across $600$ different experimental settings, focusing on model generalization and robustness. Each bold number in a row represents the best performance.
For robustness, results in \cref{tab:ra_our_cifar10} and \cref{tab:ra_our_imagenet} show that downstream models enhanced by Gen-AF are capable of accurately identifying adversarial examples with high RAs. 
For instance, in \cref{tab:ra_our_cifar10}, the robustly enhanced downstream models (based on SimCLR) identify SOTA AdvEncoder and PAP attacks with RAs of $95.47\%$ and $99.88\%$ under the full upstream-knowledge attacker setting on the GTSRB dataset, respectively. 

For generalization, by comparing the TA metrics of standard training and adversarial training with Gen-AF in \cref{tab:clean_performance} and \cref{tab:ta_our}, we observe that they are broadly similar, and in some cases, the adversarially trained models even surpass their standard counterparts. 
For instance, the SimCLR encoder pre-trained on CIFAR10 shows a downstream TA of $77.18\%$ under standard training, but an increased TA of $90.95\%$ under adversarial training. 
These results highlights the strong applicability of our proposed scheme in the  pre-training paradigm.

\subsection{Black-box Scenario} \label{sec:black-box}
In a black-box scenario where attackers have no information about the pre-trained encoder and the downstream task, our investigation focuses on transfer-based adversarial attacks.
We examine two types of transfer-based attacks based on the SOTA DAE attack AdvEncoder: cross-model attacks within the same pre-training dataset, and cross-model attacks across pre-training datasets. 
We use adversarial examples created by AdvEncoder on ten types of self-supervised pre-trained encoders across two pre-training datasets
to evaluate the model's ability against black-box attacks. 
These samples are then used in transfer-based experiments between their corresponding adversarially trained models with Gen-AF.
In \cref{fig:black_box_transfer}, each column of the subfigures represents adversarial examples crafted on the given pre-trained encoder and employed to attack downstream tasks of other pre-trained encoders. 
For instance, the first column in \cref{a} illustrates adversarial examples generated using AdvEncoder based on BYOL to attack other models like W-MSE, VibCreg, and NNCLR. 
By comparing the heat map results of \cref{a} - \cref{b} and \cref{c} - \cref{d}, it is evident that 
defending against cross-pre-training dataset transfer attacks is simpler than within the same pre-training dataset. This further emphasizes the threat posed by DAEs to the pre-training paradigm.

\begin{figure}[!t]   
  \centering
     \subcaptionbox{Module\label{fig:ab_module1}}{\vspace{-0.3em}\includegraphics[width=0.21\textwidth]
     {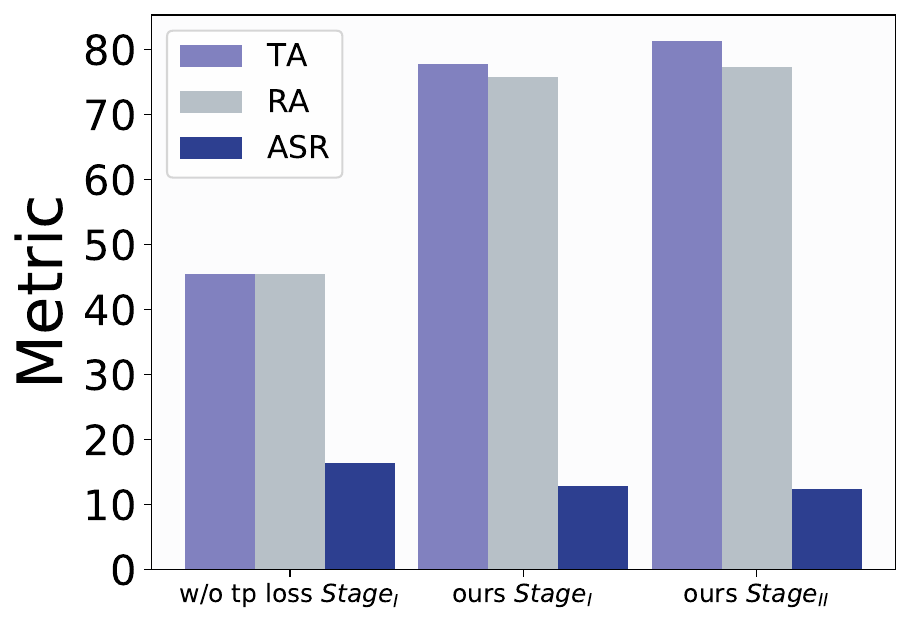}}
    \subcaptionbox{Perturbation Budget\label{fig:ab_eps}}{\vspace{-0.3em}\includegraphics[width=0.195\textwidth]{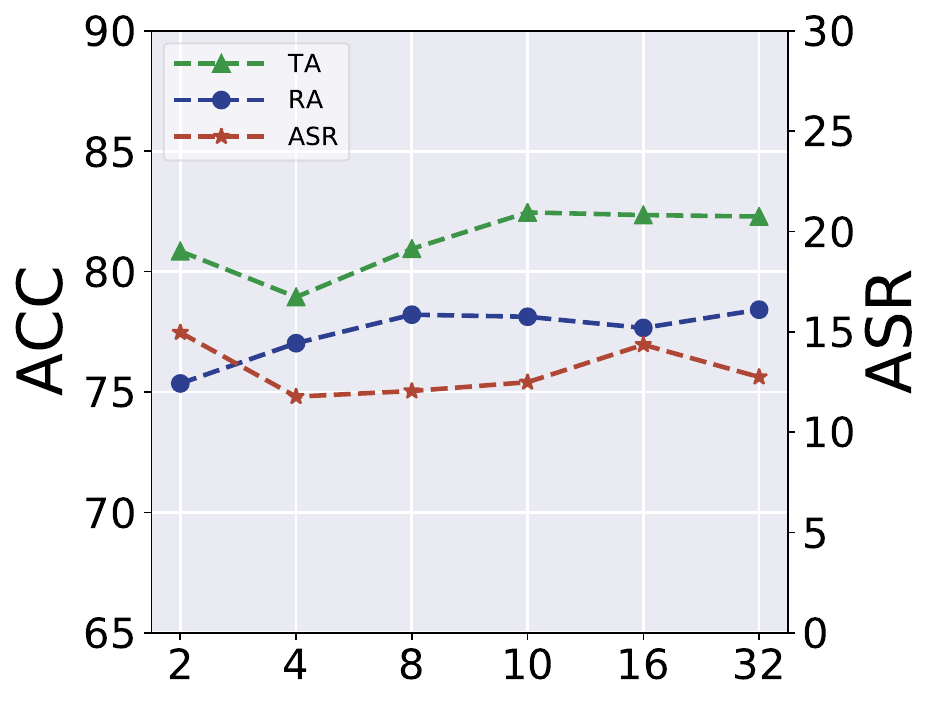}}
     \subcaptionbox{Learning Rate\label{fig:ab_lr}}{\vspace{-0.3em}\includegraphics[width=0.2\textwidth]{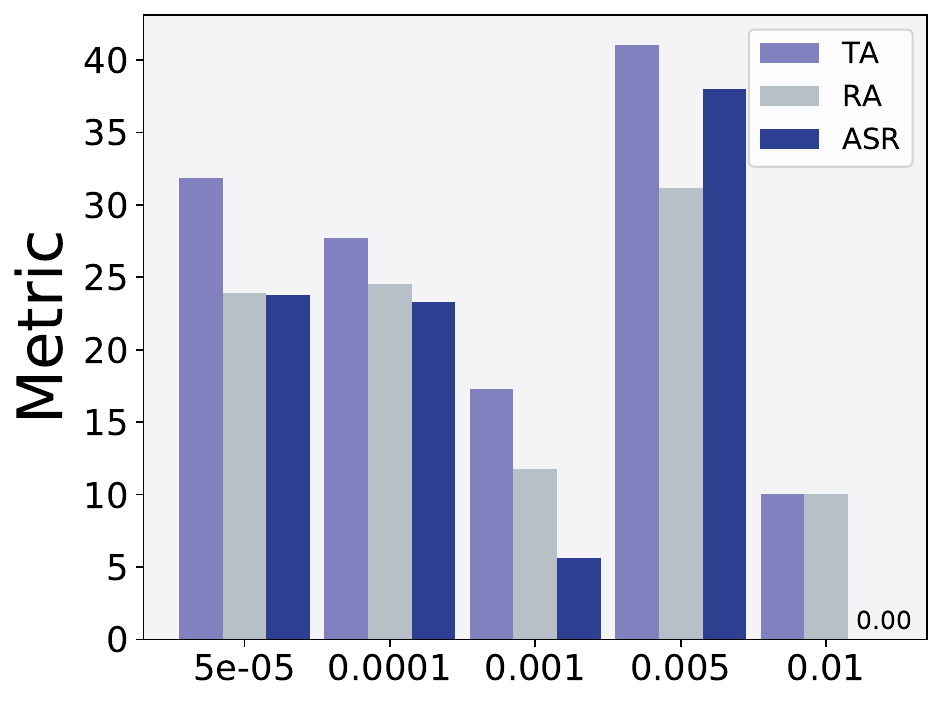}}
     \subcaptionbox{Top-k\label{fig:ab_ratio}}{\vspace{-0.3em}\includegraphics[width=0.205\textwidth]{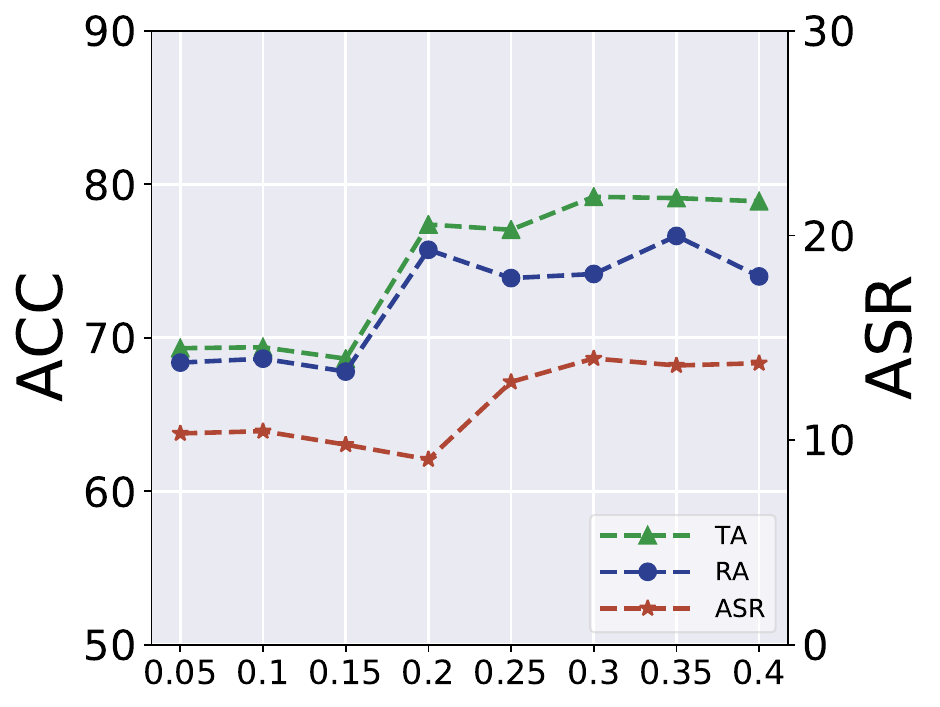}}

\caption{The ablation study under different settings  (\%) . }
\label{fig:ablation}
\end{figure}

\subsection{Ablation Study} \label{sec:ablation}
In this section, we evaluate the effect of different modules and parameters in the scheme.
We select SimCLR encoders pre-trained on CIFAR10 to conduct experiments on the STL10 dataset.

\noindent\textbf{The effect of modules.} 
We investigate the effect of  bilevel-optimizer collaborative strategy, genetic regularization loss $\mathcal{L}_{gr}$ and the strand training $\mathcal{L}_{SAT}$ in second stage on the overall scheme.
We provide the stage I results of adversarial training using a traditional single optimizer strategy in \cref{fig:ab_lr}. The results show that both the best TA and RA are below $40\%$, indicating that a single optimizer strategy is not suitable for fine-tuning pre-trained encoders. However, as shown in \cref{tab:learning_rate}, by setting different learning rates for the pre-trained encoder and classifier, denoted as $LR_E$ and $LR_C$ respectively, there is a significant improvement in the model's accuracy and robustness. This demonstrates the indispensability of the dual optimizer collaborative strategy.
As shown in \cref{fig:ab_module1}, the lack of $\mathcal{L}_{gr}$ constrains the downstream model's accuracy and robustness after the loss of pre-trained encoder knowledge. 
Based on the first stage (Stage\_I) of adversarial training, the second stage (Stage\_II) of standard training can further improve the accuracy of the model while ensuring the original robustness.

\begin{table}[t]
  \centering
  \caption{Effects of   learning rates in the bilevel-optimizer collaborative strategy.}
    \begin{tabular}{cccccc}
    \toprule[1.5pt]
   Setting &   $LR_E$  & $LR_C$  & TA $\uparrow$   & RA $\uparrow$    & ASR $\downarrow$ \\
    \midrule
   $\mathcal{S}_{1}$ & 0.01  & 0.005 & 24.51 & 22.05 & 31.52 \\
   $\mathcal{S}_{2}$ & 0.001 & 0.005 & 56.09 & 54.94 & 11.89 \\
   $\mathcal{S}_{3}$ & 0.0001 & 0.001 & 77.76 & 75.70 & 12.83 \\
    $\mathcal{S}_{4}$ & 0.0001 & 0.0001 & 45.56 & 42.00 & 22.99 \\
 Ours & 0.0001 & 0.005 & 77.84 & 75.92 & 12.71 \\
    \bottomrule[1.5pt]
    \end{tabular}%
  \label{tab:learning_rate}%
\end{table}%

\noindent\textbf{The effect of learning rates in the bilevel-optimizer collaborative strategy.} 
We further study the effect of learning rates in the bilevel-optimizer collaborative strategy on our approach. The first stage results in \cref{tab:learning_rate} indicate that setting a lower learning rate for the pre-trained encoder can achieve higher TA values in model accuracy, while a larger learning rate may lead to the collapse of the original parameters of the pre-trained model, resulting in poor performance. Conversely, a small learning rate for the classifier hinders model convergence and affects performance. These conclusions reiterate the necessity of setting different learning rates for the pre-trained encoder and classifier.

\noindent\textbf{The effect of perturbation budget.} 
We investigate how the perturbation budget $\epsilon / 255$ of adversarial examples used in training affects our method's performance. 
As shown in \cref{fig:ab_eps}, our work achieves the optimal balance between RA and TA when $\epsilon$ is set to $10$. This is attributed to our approach being more targeted towards attacks when $\epsilon$ matches the attacker's $\epsilon$ setting. However, \cref{fig:ab_eps} still indicates that even when our $\epsilon$ setting differs from that of the attackers, our approach remains effective in providing robust defense.

\noindent\textbf{The effect of top-k.} 
We investigate the effect of choosing the number of robustly insensitive layers to be fine-tuned on the final performance in the second stage. From \cref{fig:ab_ratio}, it can be seen that the TA of the model gradually increases as the ratio increases. RA and ASR have a tendency to fluctuate in this process, and generally show a decreasing trend in robustness.

\subsection{Comparison Study}\label{sec:compare}
In this section, we conduct a comparison of Gen-AF with adversarial training methods like TRADES~\cite{zhang2019theoretically}, MART~\cite{wang2019improving}, and AWP~\cite{wu2020adversarial}, where ImageNet serves as the pre-training dataset and STL10 as the downstream dataset. The experimental setup for these methods, such as perturbation budgets and epochs, aligns with ours. 
The results are provided in \cref{tab:compare}, where our method demonstrates comprehensive superiority over existing schemes in terms of both robustness and generalization.
In evaluating generalization, as illustrated in \cref{tab:clean_performance}, the SimCLR and MoCov2 models achieve TA values of  $62.81\%$ and  $64.98\%$, respectively, under standard downstream training.
Only the model trained by Gen-AF maintains or even improves its generalization, recording an increase of $6.08\%$, whereas the best TRADES decreases by $12.02\%$.
In evaluating robustness,  Gen-AF surpasses all existing schemes, maintaining a RA above $60\%$ against six types of attacks.

\begin{table}[htbp]
  \centering
  \caption{Comparison Study (\%)}
   \scalebox{0.635}{
    \begin{tabular}{cccccccc}
    \toprule[1.5pt]
    \multirow{2}[1]{*}{Model} & \multirow{2}[1]{*}{Method} & \multirow{2}[1]{*}{TA} & UAP   & UAPGD & SSP   & PAP   & AdvEncoder \\  
          &       &       & RA    & RA    & RA    & RA    & RA \\
           \midrule
    \multirow{4}[1]{*}{SimCLR} 
    & PGD-AT~\cite{carlini2017towards} & 22.19  & 21.23  & 23.00  & 21.79  & 20.97  & 19.04  \\
          & MART~\cite{wang2019improving}  & 42.79  & 42.50  & 42.65  & 42.60  & 42.24  & 42.36  \\
          & TRADES~\cite{zhang2019theoretically} & 50.79  & 50.40  & 50.51  & 50.95  & 50.59  & 50.66  \\
          & Gen-AT (Ours)   & \textbf{68.69} & \textbf{64.43}  & \textbf{68.65}  & \textbf{65.92}  & \textbf{65.48}  & \textbf{62.28}  \\
    \midrule
    \multirow{4}[2]{*}{MoCov2+} 
    & PGD-AT~\cite{carlini2017towards}  & 13.72  & 12.15  & 13.48  & 12.30  & 13.02  & 12.70 \\
          & MART~\cite{wang2019improving}  & 42.79  & 41.46  & 41.52  & 41.28  & 41.76  & 41.36  \\
          & TRADES~\cite{zhang2019theoretically}& 50.50  & 52.40  & 52.60  & 52.37  & 52.60  & 52.28  \\
          & Gen-AT(Ours)   & \textbf{69.09}  & \textbf{64.98}  & \textbf{68.15} & \textbf{59.41}  & \textbf{60.47}  & \textbf{66.00}  \\
    \bottomrule[1.5pt]
    \end{tabular}%
    }
  \label{tab:compare}%
\end{table}%
\section{Discussion }\label{sec:discussion}
In this section, we will discuss how to employ the proposed Gen-AF framework to defend adversarial patches~\cite{brown2017adversarial, zhou2023advencoder}, and backdoors~\cite{jia2022badencoder, li2022backdoor} against pre-trained encoders.

\subsection{Defense against patch-based AEs}\label{sec:discussion_patch}
Unlike \textit{adversarial perturbations}~\cite{szegedy2013intriguing, goodfellow2014explaining, carlini2017towards} that globally modify images based on $L_p$ norms, \textit{adversarial patches}~\cite{zhou2023advencoder, zhou2023advclip} make unrestricted modifications in specific areas of the image and exhibit stronger attack capabilities.
We aim to mitigate the adversarial effects of these patches by blurring their distinct features, thereby enhancing our defense against such attacks.
We propose a genetic evolution-neurtured adversarial fine-tuning method based on preprocessing (Gen-PAF), which involves first preprocessing the images and then feeding them into an adversarially trained model to achieve patch-based defense.
We select the SimCLR model pre-trained on ImageNet to test the robustness against the SOTA adversarial patch scheme AdvEncoder-Patch~\cite{zhou2023advencoder} (\ie, Adv-PAT) on the GTSRB dataset.

Initially, we preprocess data with nineteen image corrosion techniques before inputting it into a clean model. 
The visualizations of corroded patch-based adversarial examples are provided in \cref{fig:visualization} of Appendix.
In  \cref{fig:adv_patch}, the ASR of the adversarial patch significantly decreased yet maintain a certain level, with a notable drop in model accuracy.
However, when we feed the preprocessed images into the adversarially trained model with Gen-AF, the patch-based adversarial examples are correctly identified more than $90\%$,  with minimal accuracy impact.
These phenomena may be attributed to the enhanced robustness against erosion-based interference in the adversarially trained model, thus allowing normal images to be correctly recognized. Concurrently, adversarial examples bearing patches with diminished efficacy are also accurately identified.
The above results demonstrate that by incorporating a preprocessing module, we can achieve defense against adversarial patches without altering the original scheme.

\begin{figure}[!t]  
\centering
    \subcaptionbox{TA}  {\includegraphics[width=0.23\textwidth]{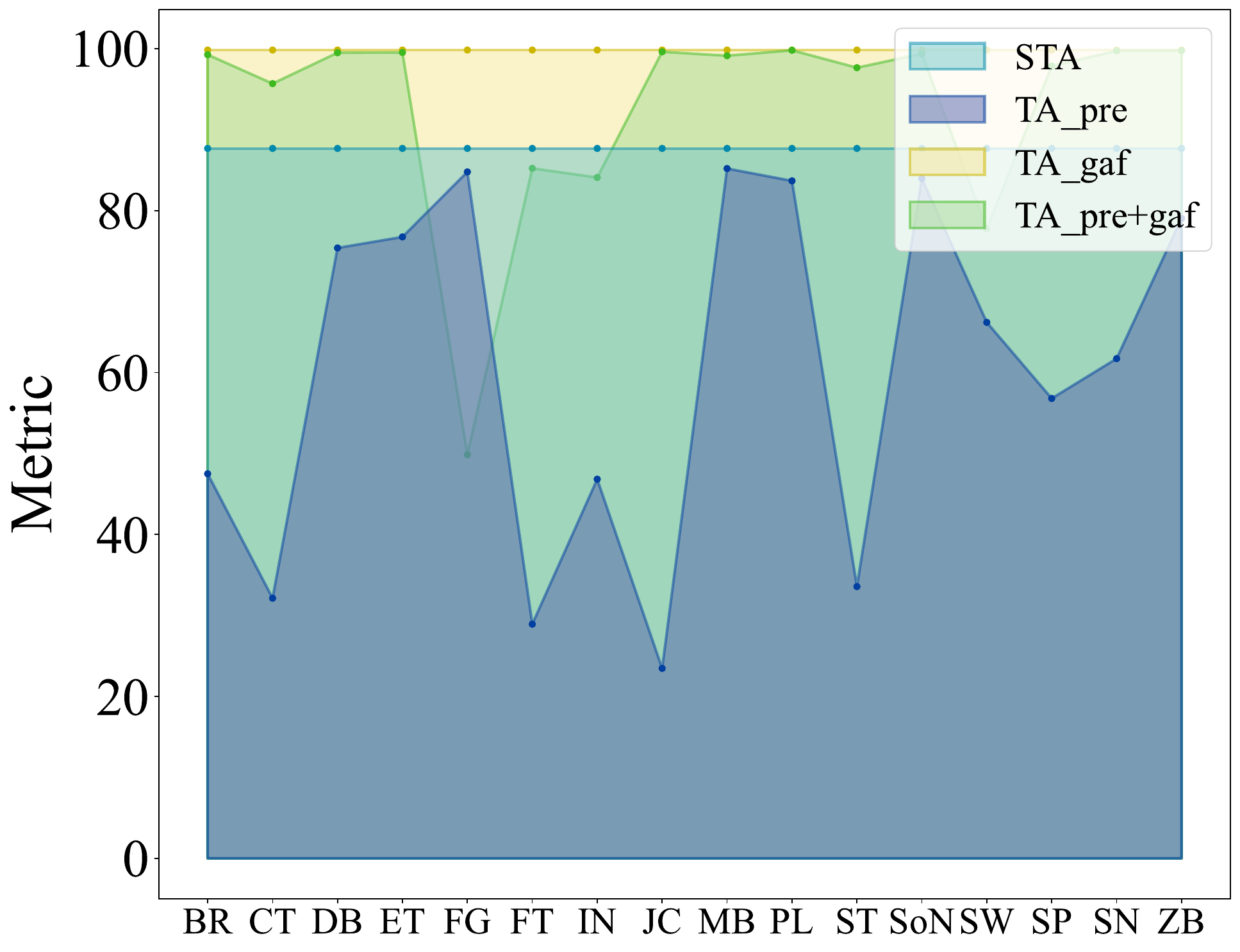}}
    \subcaptionbox{ASR}{\includegraphics[width=0.23\textwidth]{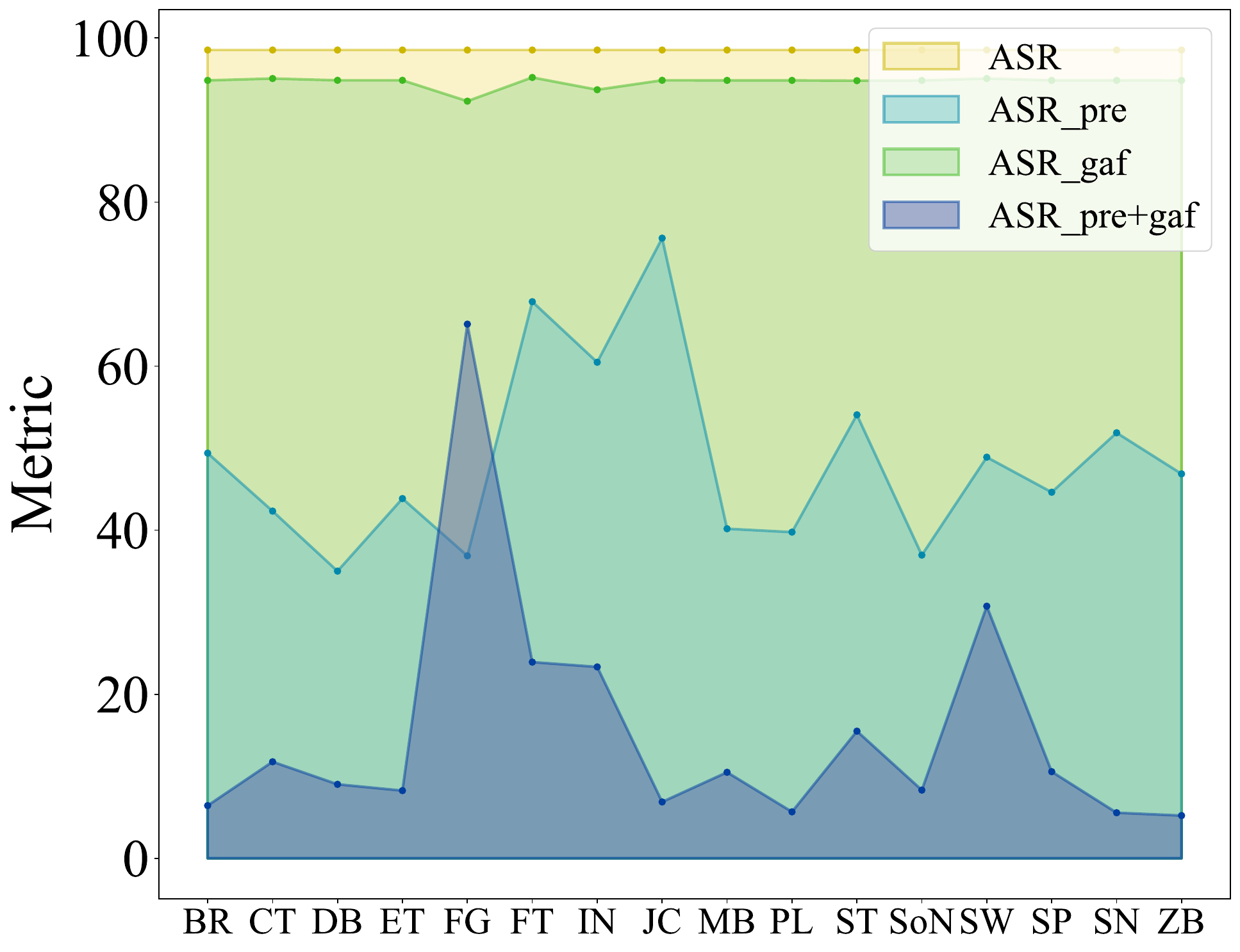}}
\caption{The results (\%) of defense against patch-based adversarial examples made with AdvEncoder. The suffix ``pre" represents the results of the preprocessing method, ``gaf" denotes the outcomes of Gen-AF, and ``pre+gaf" indicates the results of combining both.}
\label{fig:adv_patch}
 \vspace{-0.6cm}
\end{figure}

\subsection{Defense against backdoor attacks}
We explore whether our proposed method can defend \textit{backdoor} attacks when the official model is compromised with a backdoor. 
Backdoor~\cite{li2022backdoor,hu2022badhash} attacks involve embedding hidden triggers during the model training process, causing the model to exhibit predetermined incorrect behaviors when the trigger is present, while performing normally on clean samples. 
BadEncoder~\cite{jia2022badencoder} successfully embeds a backdoor into pre-trained encoders and validates it in downstream tasks. 
Based on its experimental setup, \textit{we attempt to simultaneously defend against both the backdoors and the adversarial examples using Gen-AF without any additional modifications}.
We evaluate the effectiveness of Gen-AF in this scenario using SOTA downstream-agnostic adversarial example method AdvEncoder and SOTA backdoor against pre-trained encoders method BadEncoder.

We use the backdoored SimCLR encoders which pre-trained on CIFAR10, from the BadEncoder repository and conduct adversarial training consistent with the setup in \cref{sec:experiment_setting}.
We select pre-trained encoders that are implanted with backdoors targeting specific classes in GTSRB and SVHN datasets as victim models and train downstream models using the corresponding target datasets.
A third-party malicious attacker also create DAEs based on these backdoored pre-trained encoders with CIFAR10 as the surrogate dataset. 
As seen in \cref{tab:backdoor},  our proposed approach not only eliminates backdoors but also enhances the model's robustness in correctly identifying adversarial examples. Importantly, it also improves the model's accuracy. 
We speculate that during adversarial training, the downstream model purifies the backdoor-dependent pre-trained encoder parameters, further demonstrating Gen-AF's strong applicability in downstream tasks.

\begin{table}[t]
  \centering
  \caption{Results (\%) of backdoor defense. BASR stands for the backdoor attack success rate.}
  \scalebox{0.85}{
    \begin{tabular}{cccccc}
    \toprule[1.5pt]
    \multirow{2}[2]{*}{Dataset} & \multicolumn{2}{c}{BadEncoder~\cite{jia2022badencoder}} & \multicolumn{3}{c}{Gen-AF (ours)} \\
          & TA $\uparrow$  & BASR $\downarrow$ & TA $\uparrow$  & BASR $\downarrow$  & RA $\uparrow$ \\
    \midrule
    GTSRB & 81.84 & 98.64 & 100.00 & 3.24  & 99.90 \\
    SVHN  & 58.50 & 99.14 & 93.76 & 30.74 & 93.73 \\
    \bottomrule[1.5pt]
    \end{tabular}%
    }
  \label{tab:backdoor}%
\end{table}%

\subsection{Time Overhead}
It is well known that, compared to standard training, adversarial training requires more computational overhead to enhance robustness. And existing works~\cite{carlini2017towards, zhang2019theoretically, wang2019improving} indicate that these computational costs are deemed acceptable.
Further, we compare the computational expense of our approach against TRADES. Specifically, we measure the robustness against AdvEncoder and time costs (average time for a single epoch) for models obtained by training TRADES from scratch (TRADES-ZERO) and fine-tuning pre-trained encoders (based on CIFAR10) with TRADES (TRADES-PRE) and Gen-AF on the STL10 dataset. As seen in \cref{tab:time}, our method outperforms the others on all metrics.
These results further demonstrate the viability and security of deploying Gen-AF in downstream applications.

\begin{table}[t]
  \centering
  \caption{Results (\%) of time overhead.}
   \scalebox{0.88}{
    \begin{tabular}{cccc}
    \toprule[1.5pt]
    Method & TIME (s) $\downarrow$ &  TA  $\uparrow$    & RA $\uparrow$ \\
    \midrule
    TRADES-ZERO~\cite{zhang2019theoretically} & 49.05 & 66.66 & 66.47 \\
    TRADES-PRE~\cite{zhang2019theoretically} & 60.88 & 67.37 & 67.37 \\
    Gen-AF (ours)  & \textbf{20.90} & \textbf{82.14} & \textbf{77.49} \\
    \bottomrule[1.5pt]
    \end{tabular}%
    }
  \label{tab:time}%
\end{table}%

\section{Related Work}\label{sec:related}

\subsection{Pre-training and Fine-tuning}\label{sec:finetuning}
Self-supervised learning, as an emerging paradigm in deep learning, aims to pre-train image encoders using extensive unlabeled data. These pre-trained image encoders can subsequently be utilized to build classifiers for various downstream tasks, with minimal or no labeled data required. 
Fine-tuning is a widely adopted approach to enhance the transferability of pre-trained models to downstream tasks and domain shifts. 
Typically, fine-tuning methods involve fine-tuning the last layer (linear probing) \cite{aghajanyan2021better, kumar2022finetuning} or all layers (fully fine-tuning) \cite{aghajanyan2021better, hendrycks2019using, miller21bline, kumar2022finetuning}. 

Based on \cite{tao2022exploring,fini2022self,zhou2023advencoder},
self-supervised learning schemes can be divided into the following categories: \emph{(1) contrastive learning methods}~\cite{chen2020improved,chen2021empirical,chen2020simple}), \emph{(2) negative-free methods}~\cite{grill2020bootstrap, chen2021exploring, zheng2021ressl}), \emph{(3) clustering-based methods}~\cite{caron2020unsupervised, caron2020unsupervised, caron2021emerging}, \emph{(4) Redundancy reduction-based methods}~\cite{zbontar2021barlow, ermolov2021whitening, bardes2021vicreg, lee2021vibcreg},  \emph{(5) Nearest-neighbor retrieval-based methods}~\cite{dwibedi2021little}. 
In this work, we primarily focus on image classification tasks.

\subsection{Adversarial Attacks and Defenses}
Deep neural networks (DNNs) are susceptible to adversarial examples~\cite{szegedy2013intriguing, goodfellow2014explaining, hu2023pointca, hu2022protecting}, where subtle perturbations can mislead model classification. Universal adversarial perturbation~\cite{moosavi2017universal, deng2020universal,mopuri2017fast, naseer2020self, hu2021advhash} (UAP) extends attack generality, using a single perturbation to affect multiple samples.
Adversarial attacks are classified into white-box, with full knowledge of the target model's internal structure and parameters for crafting precise attacks, and black-box, relying solely on observing the model's input and output.
With the success of pre-trained encoders, recent efforts~\cite{zhang2022towards, ban2022pre, zhou2023advencoder, zhou2023advclip} have begun to explore adversarial examples against pre-trained encoders without downstream knowledge.
Their success highlights the significant security risks inherent in the pre-training paradigm, suggesting that utilizing pre-trained encoders may be unsafe.
Existing adversarial defenses are bifurcated into data-driven~\cite{guo2017countering} methods, which cleanse samples by eradicating adversarial noise, and model-oriented~\cite{guo2017countering,zhu2017prune} approaches, which enhance the model's resilience to  adversarial inputs.
Due to domain shift between pre-training and downstream fine-tuning, and the sensitivity of pre-trained encoder parameters, existing defense methods fail to effectively mitigate DAEs.

\section{Conclusion}
In this paper, we present Gen-AF, the first genetic evolution-neurtured  adversarial fine-tuning approach, aiming to enhance downstream model robustness and generalizability. 
Our method consists of two stages. The first employs genetic-driven dual-track adversarial training with a dual-optimizer strategy and genetic regularization, reinforcing robustness without compromising pre-training knowledge. The second stage focuses on evolutionary adaptability fine-tuning, refining redundant network layers to improve generalization.
We conduct a comprehensive analysis of mainstream defense methods, underscoring their limitations in a pre-training scenario. Gen-AF demonstrates outstanding defense performance against five state-of-the-art universal adversarial attacks designed for pre-trained encoders. Evaluation spans ten popular self-supervised learning methods, two pre-training datasets, and six downstream datasets.
Experimental results showcase Gen-AF's effectiveness in defending against adversarial examples and backdoors targeting pre-trained encoders.

\section*{Acknowledgements}
This work is supported by the National Natural Science Foundation of China (Grant No.U20A20177) and Hubei Province Key R\&D Technology Special Innovation Project (Grant No.2021BAA032). 
Minghui Li and Shengshan Hu are co-corresponding authors.


{
\small
\bibliographystyle{plain}
\bibliography{sp24}

\begin{thebibliography}{10}

\bibitem{Animals10}
Animals-10-animal pictures of 10 different categories taken from google images.
\newblock Online, August 2020.

\bibitem{aghajanyan2021better}
Armen Aghajanyan, Akshat Shrivastava, Anchit Gupta, Naman Goyal, Luke Zettlemoyer, and Sonal Gupta.
\newblock Better fine-tuning by reducing representational collapse.
\newblock In {\em Proceedings of the International Conference on Learning Representations (ICLR'21)}, 2021.

\bibitem{ban2022pre}
Yuanhao Ban and Yinpeng Dong.
\newblock Pre-trained adversarial perturbations.
\newblock In {\em Proceedings of the 36th International Conference on Neural Information Processing Systems (NeurIPS'22)}, 2022.

\bibitem{bardes2021vicreg}
Adrien Bardes, Jean Ponce, and Yann LeCun.
\newblock Vicreg: Variance-invariance-covariance regularization for self-supervised learning.
\newblock {\em arXiv preprint arXiv:2105.04906}, 2021.

\bibitem{brown2017adversarial}
Tom~B. Brown, Dandelion Man{\'e}, Aurko Roy, Mart{\'\i}n Abadi, and Justin Gilmer.
\newblock Adversarial patch.
\newblock {\em arXiv preprint arXiv:1712.09665}, 2017.

\bibitem{brown2020language}
Tom~B. Brown, Benjamin Mann, Nick Ryder, Melanie Subbiah, Jared Kaplan, Prafulla Dhariwal, Arvind Neelakantan, Pranav Shyam, Girish Sastry, Amanda Askell, Sandhini Agarwal, Ariel Herbert{-}Voss, Gretchen Krueger, Tom Henighan, Rewon Child, Aditya Ramesh, Daniel~M. Ziegler, Jeffrey Wu, Clemens Winter, Christopher Hesse, Mark Chen, Eric Sigler, Mateusz Litwin, Scott Gray, Benjamin Chess, Jack Clark, Christopher Berner, Sam McCandlish, Alec Radford, Ilya Sutskever, and Dario Amodei.
\newblock Language models are few-shot learners.
\newblock In {\em Proceedings of the 34th International Conference on Neural Information Processing Systems (NeurIPS'20)}, pages 1877--1901, 2020.

\bibitem{carlini2017towards}
Nicholas Carlini and David Wagner.
\newblock Towards evaluating the robustness of neural networks.
\newblock In {\em Proceedings of the 38th IEEE Symposium on Security and Privacy (S\&P'17)}, pages 39--57. Ieee, 2017.

\bibitem{caron2020unsupervised}
Mathilde Caron, Ishan Misra, Julien Mairal, Priya Goyal, Piotr Bojanowski, and Armand Joulin.
\newblock Unsupervised learning of visual features by contrasting cluster assignments.
\newblock In {\em Proceedings of the 34th International Conference on Neural Information Processing Systems (NeurIPS'20)}, pages 9912--9924, 2020.

\bibitem{caron2021emerging}
Mathilde Caron, Hugo Touvron, Ishan Misra, Herv{\'e} J{\'e}gou, Julien Mairal, Piotr Bojanowski, and Armand Joulin.
\newblock Emerging properties in self-supervised vision transformers.
\newblock In {\em Proceedings of the IEEE/CVF International Conference on Computer Vision (ICCV'21)}, pages 9650--9660, 2021.

\bibitem{chatterji2019intriguing}
Niladri~S Chatterji, Behnam Neyshabur, and Hanie Sedghi.
\newblock The intriguing role of module criticality in the generalization of deep networks.
\newblock In {\em Proceedings of the International Conference on Learning Representations (ICLR'20)}, 2020.

\bibitem{chen2020robust}
Tianlong Chen, Zhenyu Zhang, Sijia Liu, Shiyu Chang, and Zhangyang Wang.
\newblock Robust overfitting may be mitigated by properly learned smoothening.
\newblock In {\em International Conference on Learning Representations}, 2020.

\bibitem{chen2020simple}
Ting Chen, Simon Kornblith, Mohammad Norouzi, and Geoffrey Hinton.
\newblock A simple framework for contrastive learning of visual representations.
\newblock In {\em Proceedings of the International Conference on Machine Learning (ICML'20)}, pages 1597--1607. PMLR, 2020.

\bibitem{chen2020improved}
Xinlei Chen, Haoqi Fan, Ross Girshick, and Kaiming He.
\newblock Improved baselines with momentum contrastive learning.
\newblock {\em arXiv preprint arXiv:2003.04297}, 2020.

\bibitem{chen2021exploring}
Xinlei Chen and Kaiming He.
\newblock Exploring simple siamese representation learning.
\newblock In {\em Proceedings of the IEEE/CVF Conference on Computer Vision and Pattern Recognition (CVPR'21)}, pages 15750--15758, 2021.

\bibitem{chen2021empirical}
Xinlei Chen, Saining Xie, and Kaiming He.
\newblock An empirical study of training self-supervised vision transformers.
\newblock In {\em Proceedings of the IEEE/CVF International Conference on Computer Vision (ICCV'21)}, pages 9640--9649, 2021.

\bibitem{coates2011analysis}
Adam Coates, Andrew Ng, and Honglak Lee.
\newblock An analysis of single-layer networks in unsupervised feature learning.
\newblock In {\em Proceedings of the Fourteenth International Conference on Artificial Intelligence and Statistics (AISTATS'11)}, pages 215--223. JMLR Workshop and Conference Proceedings, 2011.

\bibitem{deng2020universal}
Yingpeng Deng and Lina~J. Karam.
\newblock Universal adversarial attack via enhanced projected gradient descent.
\newblock In {\em Proceedings of the IEEE International Conference on Image Processing (ICIP'20)}, pages 1241--1245. IEEE, 2020.

\bibitem{dwibedi2021little}
Debidatta Dwibedi, Yusuf Aytar, Jonathan Tompson, Pierre Sermanet, and Andrew Zisserman.
\newblock With a little help from my friends: Nearest-neighbor contrastive learning of visual representations.
\newblock In {\em Proceedings of the IEEE/CVF International Conference on Computer Vision (ICCV'21)}, pages 9588--9597, 2021.

\bibitem{ermolov2021whitening}
Aleksandr Ermolov, Aliaksandr Siarohin, Enver Sangineto, and Nicu Sebe.
\newblock Whitening for self-supervised representation learning.
\newblock In {\em Proceedings of the International Conference on Machine Learning (ICML'21)}, pages 3015--3024. PMLR, 2021.

\bibitem{fini2022self}
Enrico Fini, Victor G.~Turrisi da~Costa, Xavier Alameda-Pineda, Elisa Ricci, Karteek Alahari, and Julien Mairal.
\newblock Self-supervised models are continual learners.
\newblock In {\em Proceedings of the IEEE/CVF Conference on Computer Vision and Pattern Recognition (CVPR'22)}, pages 9621--9630, 2022.

\bibitem{goodfellow2014explaining}
Ian~J Goodfellow, Jonathon Shlens, and Christian Szegedy.
\newblock Explaining and harnessing adversarial examples.
\newblock {\em arXiv preprint arXiv:1412.6572}, 2014.

\bibitem{grill2020bootstrap}
Jean{-}Bastien Grill, Florian Strub, Florent Altch{\'{e}}, Corentin Tallec, Pierre~H. Richemond, Elena Buchatskaya, Carl Doersch, Bernardo~{\'{A}}vila Pires, Zhaohan Guo, Mohammad~Gheshlaghi Azar, Bilal Piot, Koray Kavukcuoglu, R{\'{e}}mi Munos, and Michal Valko.
\newblock Bootstrap your own latent a new approach to self-supervised learning.
\newblock In {\em Proceedings of the 34th International Conference on Neural Information Processing Systems (NeurIPS'20)}, pages 21271--21284, 2020.

\bibitem{guo2017countering}
Chuan Guo, Mayank Rana, Moustapha Cisse, and Laurens Van Der~Maaten.
\newblock Countering adversarial images using input transformations.
\newblock {\em arXiv preprint arXiv:1711.00117}, 2017.

\bibitem{hendrycks2019using}
Dan Hendrycks, Kimin Lee, and Mantas Mazeika.
\newblock Using pre-training can improve model robustness and uncertainty.
\newblock In {\em Proceedings of the 36th International Conference on Machine Learning (ICML'19)}, pages 2712--2721. PMLR, 2019.

\bibitem{hu2022protecting}
Shengshan Hu, Xiaogeng Liu, Yechao Zhang, Minghui Li, Leo~Yu Zhang, Hai Jin, and Libing Wu.
\newblock Protecting facial privacy: Generating adversarial identity masks via style-robust makeup transfer.
\newblock In {\em Proceedings of the IEEE/CVF Conference on Computer Vision and Pattern Recognition (CVPR'22)}, pages 15014--15023, 2022.

\bibitem{hu2023pointca}
Shengshan Hu, Junwei Zhang, Wei Liu, Junhui Hou, Minghui Li, Leo~Yu Zhang, Hai Jin, and Lichao Sun.
\newblock Pointca: Evaluating the robustness of 3d point cloud completion models against adversarial examples.
\newblock In {\em Proceedings of the 37th AAAI Conference on Artificial Intelligence (AAAI'23)}, number~1, pages 872--880, 2023.

\bibitem{hu2021advhash}
Shengshan Hu, Yechao Zhang, Xiaogeng Liu, Leo~Yu Zhang, Minghui Li, and Hai Jin.
\newblock Advhash: Set-to-set targeted attack on deep hashing with one single adversarial patch.
\newblock In {\em Proceedings of the 29th ACM International Conference on Multimedia (ACM MM'21)}, pages 2335--2343, 2021.

\bibitem{hu2022badhash}
Shengshan Hu, Ziqi Zhou, Yechao Zhang, Leo~Yu Zhang, Yifeng Zheng, Yuanyuan He, and Hai Jin.
\newblock Badhash: Invisible backdoor attacks against deep hashing with clean label.
\newblock In {\em Proceedings of the 30th ACM International Conference on Multimedia (ACM MM'22)}, pages 678--686, 2022.

\bibitem{huang2023clip2point}
Tianyu Huang, Bowen Dong, Yunhan Yang, Xiaoshui Huang, Rynson~WH Lau, Wanli Ouyang, and Wangmeng Zuo.
\newblock Clip2point: Transfer clip to point cloud classification with image-depth pre-training.
\newblock In {\em Proceedings of the IEEE/CVF International Conference on Computer Vision (ICCV'23)}, pages 22157--22167, 2023.

\bibitem{jia2022badencoder}
Jinyuan Jia, Yupei Liu, and Neil~Zhenqiang Gong.
\newblock Badencoder: Backdoor attacks to pre-trained encoders in self-supervised learning.
\newblock In {\em Proceedings of the IEEE Symposium on Security and Privacy (S\&P'22)}, pages 2043--2059. IEEE, 2022.

\bibitem{krizhevsky2009learning}
Alex Krizhevsky and Geoffrey Hinton.
\newblock Learning multiple layers of features from tiny images.
\newblock 2009.

\bibitem{kumar2022finetuning}
Ananya Kumar, Aditi Raghunathan, Robbie~Matthew Jones, Tengyu Ma, and Percy Liang.
\newblock Fine-tuning can distort pretrained features and underperform out-of-distribution.
\newblock In {\em Proceedings of the International Conference on Learning Representations (ICLR'22)}, 2022.

\bibitem{lee2021vibcreg}
Daesoo Lee and Erlend Aune.
\newblock Vibcreg: Variance-invariance-better-covariance regularization for self-supervised learning on time series.
\newblock {\em arXiv preprint arXiv:2109.00783}, 2021.

\bibitem{li2022backdoor}
Yiming Li, Yong Jiang, Zhifeng Li, and Shu-Tao Xia.
\newblock Backdoor learning: A survey.
\newblock {\em IEEE Transactions on Neural Networks and Learning Systems}, 2022.

\bibitem{liang2018detecting}
Bin Liang, Hongcheng Li, Miaoqiang Su, Xirong Li, Wenchang Shi, and Xiaofeng Wang.
\newblock Detecting adversarial image examples in deep neural networks with adaptive noise reduction.
\newblock {\em IEEE Transactions on Dependable and Secure Computing}, 18(1):72--85, 2018.

\bibitem{liu2022poisonedencoder}
Hongbin Liu, Jinyuan Jia, and Neil~Zhenqiang Gong.
\newblock Poisonedencoder: Poisoning the unlabeled pre-training data in contrastive learning.
\newblock In {\em Proceedings of the 31st USENIX Security Symposium (USENIX Security'22)}, pages 3629--3645, 2022.

\bibitem{mi2023topology}
Xiaoyue Mi, Fan Tang, Yepeng Weng, Danding Wang, Juan Cao, Sheng Tang, Peng Li, and Yang Liu.
\newblock Topology-preserving adversarial training.
\newblock {\em arXiv preprint arXiv:2311.17607}, 2023.

\bibitem{miller21bline}
John~P Miller, Rohan Taori, Aditi Raghunathan, Shiori Sagawa, Pang~Wei Koh, Vaishaal Shankar, Percy Liang, Yair Carmon, and Ludwig Schmidt.
\newblock Accuracy on the line: on the strong correlation between out-of-distribution and in-distribution generalization.
\newblock In {\em Proceedings of the 38th International Conference on Machine Learning (ICML'21)}, pages 7721--7735, 2021.

\bibitem{moosavi2017universal}
Seyed-Mohsen Moosavi-Dezfooli, Alhussein Fawzi, Omar Fawzi, and Pascal Frossard.
\newblock Universal adversarial perturbations.
\newblock In {\em Proceedings of the IEEE/CVF Conference on Computer Vision and Pattern Recognition (CVPR'17)}, pages 1765--1773, 2017.

\bibitem{mopuri2017fast}
Konda~Reddy Mopuri, Utsav Garg, and Venkatesh~Babu Radhakrishnan.
\newblock Fast feature fool: {A} data independent approach to universal adversarial perturbations.
\newblock In {\em Proceedings of the British Machine Vision Conference (BMVC'17)}, 2017.

\bibitem{naseer2020self}
Muzammal Naseer, Salman Khan, Munawar Hayat, Fahad~Shahbaz Khan, and Fatih Porikli.
\newblock A self-supervised approach for adversarial robustness.
\newblock In {\em Proceedings of the IEEE/CVF Conference on Computer Vision and Pattern Recognition (CVPR'20)}, pages 262--271, 2020.

\bibitem{netzer2011reading}
Yuval Netzer, Tao Wang, Adam Coates, Alessandro Bissacco, Bo~Wu, and Andrew~Y Ng.
\newblock Reading digits in natural images with unsupervised feature learning.
\newblock In {\em NIPS Workshop on Deep Learning and Unsupervised Feature Learning}, 2011.

\bibitem{nie2022diffusion}
Weili Nie, Brandon Guo, Yujia Huang, Chaowei Xiao, Arash Vahdat, and Animashree Anandkumar.
\newblock Diffusion models for adversarial purification.
\newblock In {\em International Conference on Machine Learning}, pages 16805--16827. PMLR, 2022.

\bibitem{papernot2016distillation}
Nicolas Papernot, Patrick McDaniel, Xi~Wu, Somesh Jha, and Ananthram Swami.
\newblock Distillation as a defense to adversarial perturbations against deep neural networks.
\newblock In {\em Proceedings of the IEEE Symposium on Security and Privacy (SP'16)}, pages 582--597. IEEE, 2016.

\bibitem{radford2021learning}
Alec Radford, Jong~Wook Kim, Chris Hallacy, Aditya Ramesh, Gabriel Goh, Sandhini Agarwal, Girish Sastry, Amanda Askell, Pamela Mishkin, Jack Clark, et~al.
\newblock Learning transferable visual models from natural language supervision.
\newblock In {\em Proceedings of the International Conference on Machine Learning (ICML'21)}, pages 8748--8763. PMLR, 2021.

\bibitem{rasheed2023fine}
Hanoona Rasheed, Muhammad~Uzair Khattak, Muhammad Maaz, Salman Khan, and Fahad~Shahbaz Khan.
\newblock Fine-tuned clip models are efficient video learners.
\newblock In {\em Proceedings of the IEEE/CVF Conference on Computer Vision and Pattern Recognition (CVPR'23)}, pages 6545--6554, 2023.

\bibitem{rice2020overfitting}
Leslie Rice, Eric Wong, and Zico Kolter.
\newblock Overfitting in adversarially robust deep learning.
\newblock In {\em International Conference on Machine Learning}, pages 8093--8104. PMLR, 2020.

\bibitem{russakovsky2015ImageNet}
Olga Russakovsky, Jia Deng, Hao Su, Jonathan Krause, Sanjeev Satheesh, Sean Ma, Zhiheng Huang, Andrej Karpathy, Aditya Khosla, Michael~S. Bernstein, Alexander~C. Berg, and Li~Feifei.
\newblock Imagenet large scale visual recognition challenge.
\newblock {\em International Journal of Computer Vision}, 115(3):211--252, 2015.

\bibitem{saha2022backdoor}
Aniruddha Saha, Ajinkya Tejankar, Soroush~Abbasi Koohpayegani, and Hamed Pirsiavash.
\newblock Backdoor attacks on self-supervised learning.
\newblock In {\em Proceedings of the IEEE/CVF Conference on Computer Vision and Pattern Recognition (CVPR'22)}, pages 13337--13346, 2022.

\bibitem{stallkamp2012man}
Johannes Stallkamp, Marc Schlipsing, Jan Salmen, and Christian Igel.
\newblock Man vs. computer: Benchmarking machine learning algorithms for traffic sign recognition.
\newblock {\em Neural Networks}, 32:323--332, 2012.

\bibitem{szegedy2013intriguing}
Christian Szegedy, Wojciech Zaremba, Ilya Sutskever, Joan Bruna, Dumitru Erhan, Ian Goodfellow, and Rob Fergus.
\newblock Intriguing properties of neural networks.
\newblock {\em arXiv preprint arXiv:1312.6199}, 2013.

\bibitem{tao2022exploring}
Chenxin Tao, Honghui Wang, Xizhou Zhu, Jiahua Dong, Shiji Song, Gao Huang, and Jifeng Dai.
\newblock Exploring the equivalence of siamese self-supervised learning via a unified gradient framework.
\newblock In {\em Proceedings of the IEEE/CVF Conference on Computer Vision and Pattern Recognition (CVPR'22)}, pages 14431--14440, 2022.

\bibitem{tao2018attacks}
Guanhong Tao, Shiqing Ma, Yingqi Liu, and Xiangyu Zhang.
\newblock Attacks meet interpretability: Attribute-steered detection of adversarial samples.
\newblock In {\em Proceedings of the 32nd International Conference on Neural Information Processing Systems (NeurIPS'18)}, 2018.

\bibitem{wang2023corrupting}
Xianlong Wang, Shengshan Hu, Minghui Li, Zhifei Yu, Ziqi Zhou, Leo~Yu Zhang, and Hai Jin.
\newblock Corrupting convolution-based unlearnable datasets with pixel-based image transformations.
\newblock {\em arXiv preprint arXiv:2311.18403}, 2023.

\bibitem{wang2019improving}
Yisen Wang, Difan Zou, Jinfeng Yi, James Bailey, Xingjun Ma, and Quanquan Gu.
\newblock Improving adversarial robustness requires revisiting misclassified examples.
\newblock In {\em Proceedings of the International Conference on Learning Representations (ICLR'19)}, 2019.

\bibitem{wu2020adversarial}
Dongxian Wu, Shu-Tao Xia, and Yisen Wang.
\newblock Adversarial weight perturbation helps robust generalization.
\newblock {\em Proceedings of the Advances in Neural Information Processing Systems (NeurIPS'20)}, 33:2958--2969, 2020.

\bibitem{yin2019gat}
Xuwang Yin, Soheil Kolouri, and Gustavo~K Rohde.
\newblock Gat: Generative adversarial training for adversarial example detection and robust classification.
\newblock In {\em Proceedings of the International Conference on Learning Representations (ICLR'19)}, 2019.

\bibitem{zbontar2021barlow}
Jure Zbontar, Li~Jing, Ishan Misra, Yann LeCun, and St{\'e}phane Deny.
\newblock Barlow twins: Self-supervised learning via redundancy reduction.
\newblock In {\em Proceedings of the International Conference on Machine Learning (ICML'21)}, pages 12310--12320. PMLR, 2021.

\bibitem{zhang2019theoretically}
Hongyang Zhang, Yaodong Yu, Jiantao Jiao, Eric Xing, Laurent El~Ghaoui, and Michael Jordan.
\newblock Theoretically principled trade-off between robustness and accuracy.
\newblock In {\em Proceedings of the International Conference on Machine Learning (ICML'19)}, pages 7472--7482. PMLR, 2019.

\bibitem{zhang2022towards}
Jiaming Zhang, Qi~Yi, and Jitao Sang.
\newblock Towards adversarial attack on vision-language pre-training models.
\newblock In {\em Proceedings of the 30th ACM International Conference on Multimedia (ACM MM'22)}, pages 5005--5013, 2022.

\bibitem{zhang2024whydoes}
Yechao Zhang, Shengshan Hu, Leo~Yu Zhang, Junyu Shi, Minghui Li, Xiaogeng Liu, and Hai Jin.
\newblock {Why Does Little Robustness Help? A Further Step Towards Understanding Adversarial Transferability}.
\newblock In {\em Proceedings of the 45th IEEE Symposium on Security and Privacy (S\&P'24)}, 2024.

\bibitem{zheng2021ressl}
Mingkai Zheng, Shan You, Fei Wang, Chen Qian, Changshui Zhang, Xiaogang Wang, and Chang Xu.
\newblock Ressl: Relational self-supervised learning with weak augmentation.
\newblock In {\em Proceedings of the 35th International Conference on Neural Information Processing Systems (NeurIPS'21)}, pages 2543--2555, 2021.

\bibitem{zhou2023advclip}
Ziqi Zhou, Shengshan Hu, Minghui Li, Hangtao Zhang, Yechao Zhang, and Hai Jin.
\newblock Advclip: Downstream-agnostic adversarial examples in multimodal contrastive learning.
\newblock In {\em Proceedings of the 31st ACM International Conference on Multimedia (ACM MM'23)}, pages 6311--6320, 2023.

\bibitem{zhou2023advencoder}
Ziqi Zhou, Shengshan Hu, Ruizhi Zhao, Qian Wang, Leo~Yu Zhang, Junhui Hou, and Hai Jin.
\newblock Downstream-agnostic adversarial examples.
\newblock In {\em Proceedings of the IEEE/CVF International Conference on Computer Vision (ICCV'23)}, pages 4345--4355, 2023.

\bibitem{zhou2023zegclip}
Ziqin Zhou, Yinjie Lei, Bowen Zhang, Lingqiao Liu, and Yifan Liu.
\newblock Zegclip: Towards adapting clip for zero-shot semantic segmentation.
\newblock In {\em Proceedings of the IEEE/CVF Conference on Computer Vision and Pattern Recognition (CVPR'23)}, pages 11175--11185, 2023.

\bibitem{zhu2023improving}
Kaijie Zhu, Xixu Hu, Jindong Wang, Xing Xie, and Ge~Yang.
\newblock Improving generalization of adversarial training via robust critical fine-tuning.
\newblock In {\em Proceedings of the IEEE/CVF International Conference on Computer Vision (ICCV'23)}, pages 4424--4434, 2023.

\bibitem{zhu2017prune}
Michael Zhu and Suyog Gupta.
\newblock To prune, or not to prune: exploring the efficacy of pruning for model compression.
\newblock {\em arXiv preprint arXiv:1710.01878}, 2017.

\end{thebibliography}
}
\appendices\label{Appendix}
\section{Appendix Contents}
\setcounter{table}{0}
\setcounter{figure}{0}
\renewcommand{\thetable}{A\arabic{table}}
\renewcommand{\thefigure}{A\arabic{figure}}

\subsection{Datasets}
 We use the following six image datasets. For computational efficiency, we resize each image to 64$\times$64$\times$3.
\begin{itemize}
    
\item {\bf CIFAR10~\cite{krizhevsky2009learning}:} This dataset contains 50,000  training images and 10,000  testing images. Each image has a size of 32$\times$32$\times$3 and belongs to one of 10 classes. 

\item {\bf STL10~\cite{coates2011analysis}:} 
This dataset contains 5,000 labeled training images and 8,000 labeled testing images, each of which has a size of 96$\times$96$\times$3. Moreover, the dataset contains 10 classes and each image belongs to one of them. Besides the labeled training and testing images, the dataset also contains 100,000 unlabeled images.  

\item {\bf GTSRB~\cite{stallkamp2012man}:} This dataset contains 51,800 traffic sign images in 43 categories. Each image has a size of 32$\times$32$\times$3. The dataset is divided into 39,200 training images and 12,600 testing images. 

\item {\bf ImageNet~\cite{russakovsky2015ImageNet}:} This dataset contains $1.2M$ training samples and $50,000$ testing samples with $1000$ classes. Each image has a size of 256×256×3. We randomly select $20$ classes from ImageNet to build downstream dataset, , denoted as ImageNet20.

\item {\bf SVHN~\cite{netzer2011reading}:} 
This dataset contains 73,257 training images and 26,032 testing images. The size of each image is 32$\times$32$\times$3. Moreover, each image belongs to one of the 10 digits. 

\item {\bf Animals10~\cite{Animals10}:} This dataset comprises 10,000 training images and 2,500 testing images. Each image is of dimensions 64$\times$64$\times$3 and belongs to one of 10 distinct animal classes.
\end{itemize}

 \begin{figure*}[!t]
    \centering
    \includegraphics[scale=0.455]{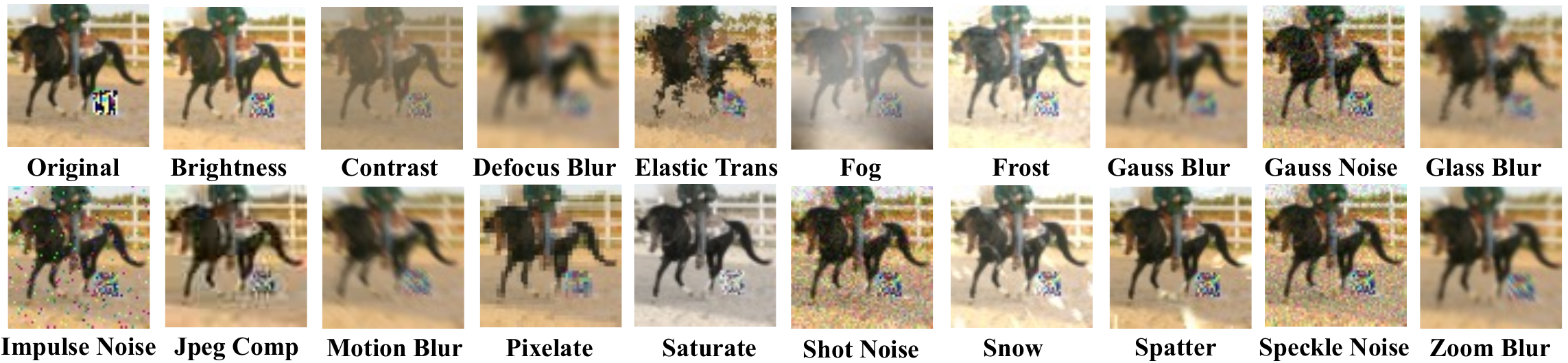}
    \caption{Visualization of corroded patch-based adversarial examples, which are produced by AdvEncoder.}
    \label{fig:visualization}
\end{figure*}

\begin{figure}[!t]   
  \centering
 \subcaptionbox{Surrogate Dataset \label{fig:ab_surrogagte_dataset}}{\includegraphics[width=0.22\textwidth]{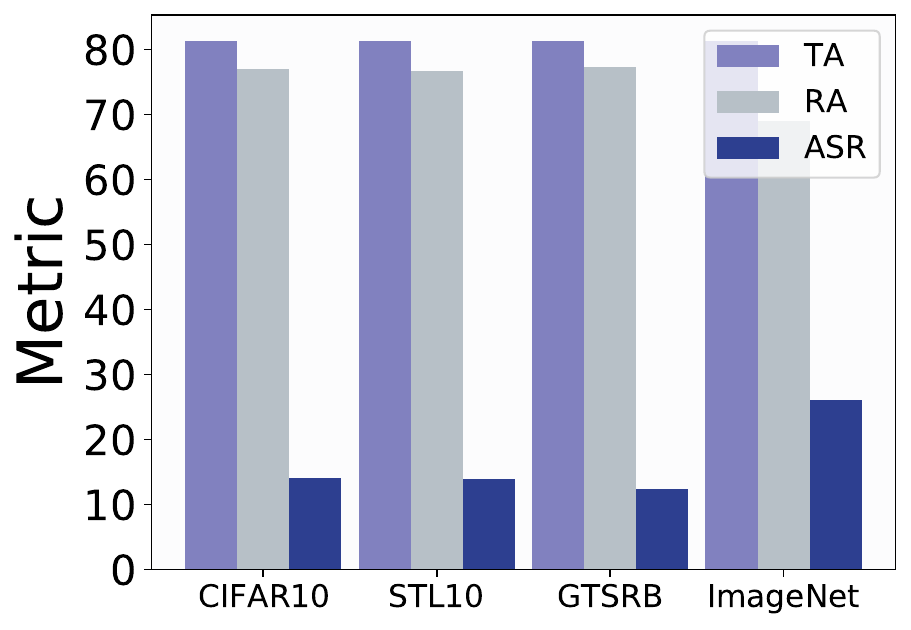}}
\subcaptionbox{Nums \label{fig:ab_dataset_nums}}{\includegraphics[width=0.21\textwidth]{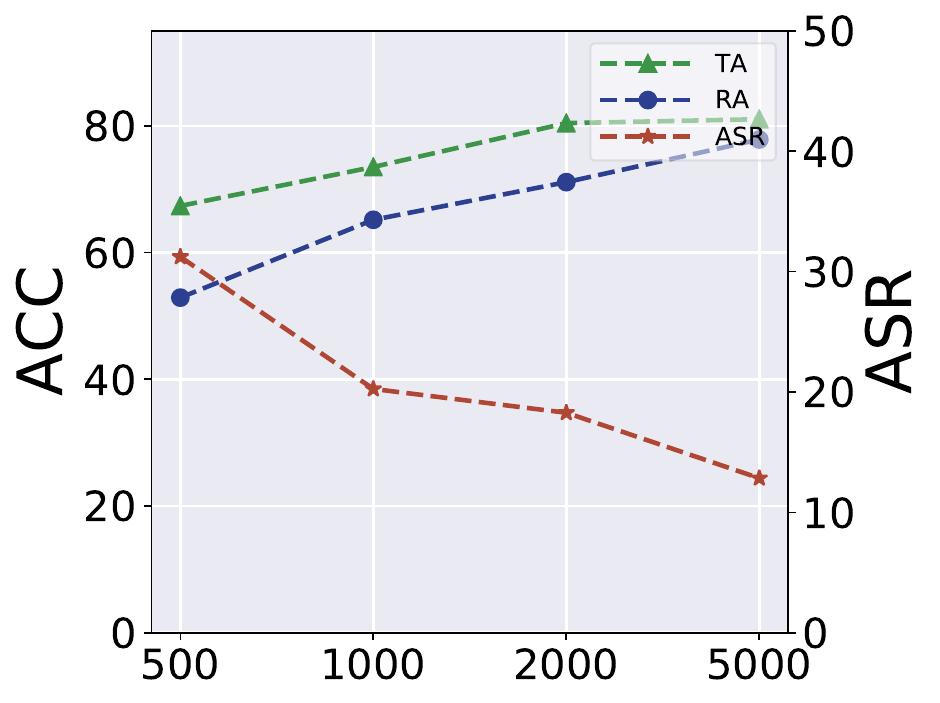}}
  \subcaptionbox{ResNet50-GTSRB \label{fig:r50_GTSRB}}{\includegraphics[width=0.22\textwidth]{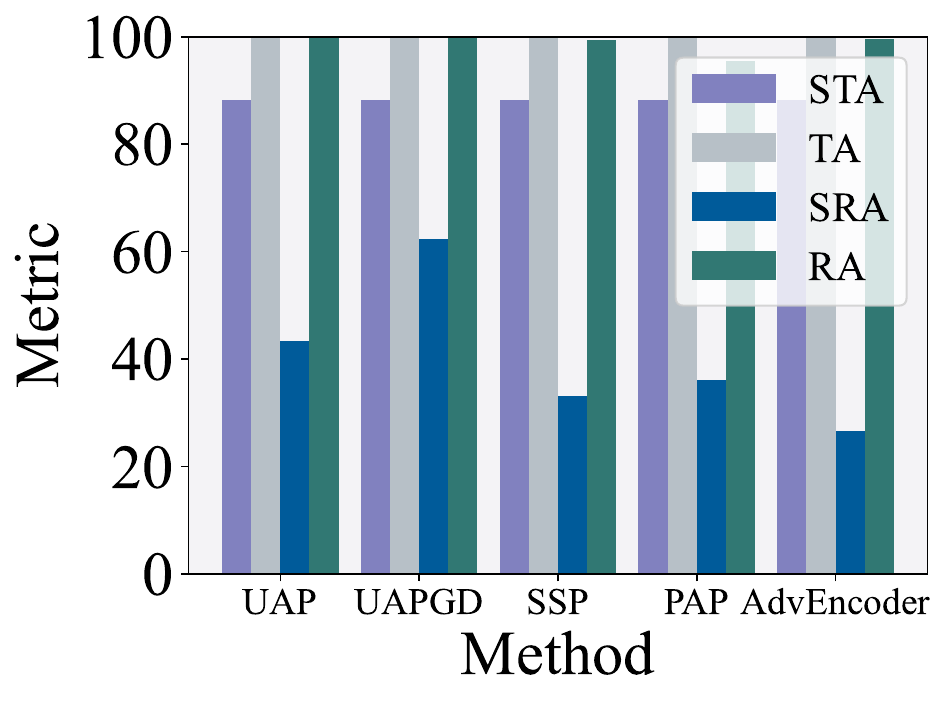}}
     \subcaptionbox{ResNet50-STL10 \label{fig:r50_STL10}}{\includegraphics[width=0.22\textwidth]{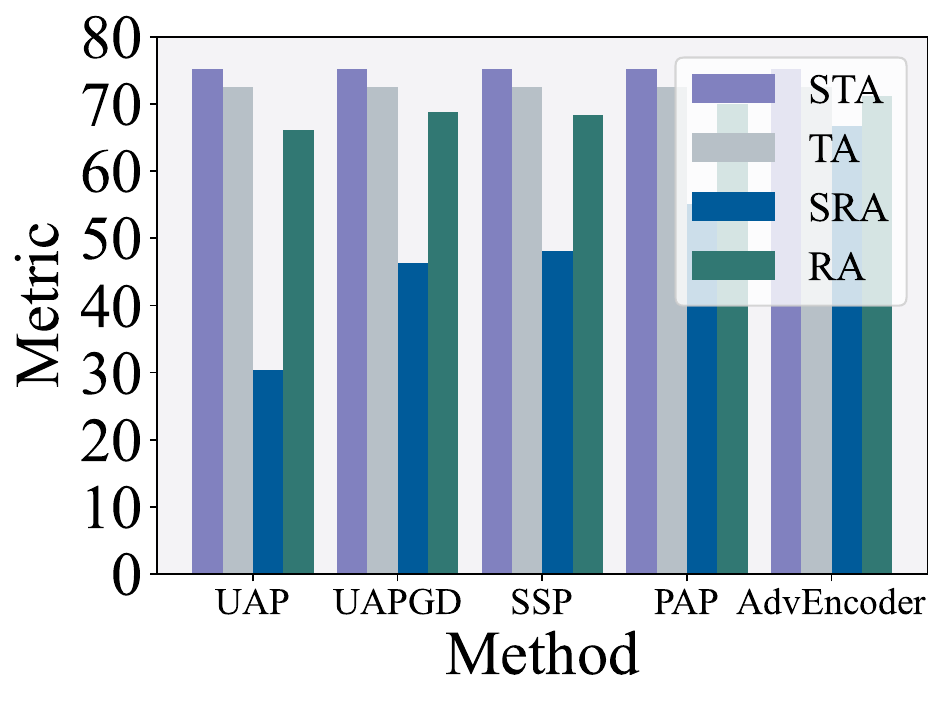}}
\caption{Effect (\%)  of other experimental settings. }
\label{fig:ablation_appendix}
\end{figure}

\subsection{Attack Performance}
We present the attack performance of UAP, UAPGD, SSP, PAP, and AdvEncoder under the same settings as in \cref{sec:qbox}, detailed in \cref{tab:attack}. 
Under the full upstream-knowledge attacker setting, PAP exhibits the best overall performance, while AdvEncoder performs optimally in the partial upstream-knowledge attacker setting. These results indicate that downstream-agnostic adversarial examples designed based on pre-trained encoders demonstrate excellent cross-domain attack capabilities.

\subsection{Visualization}
We provide visualizations of patch-based adversarial examples created using AdvEncoder, corroded by nineteen preprocessing methods~\cite{wang2023corrupting} at severity level 1 as described in \cref{sec:discussion_patch}. 
The visualization results show that the adversarial patches are noticeably blurred.

\subsection{Effect of Other Experimental Settings}
We explore the effect of different settings on our approach using the same experimental setup as described in \cref{sec:ablation}.

\noindent\textbf{The effect of attacker's surrogate dataset.} 
Considering the scenario where the attacker can employ various surrogate datasets, we evaluate the performance of our approach when under attack from four attacker's surrogate datasets. 
{From \cref{fig:ab_surrogagte_dataset}}, it is evident that regardless of the surrogate dataset utilized by the attacker to generate adversarial examples, there is minimal impact on the ASR against the downstream task. The model's RA remains consistently high.

\noindent\textbf{The effect of backbone.} 
To explore the impact of different backbones on Gen-AT, as shown in the \cref{fig:r50_GTSRB} and \cref{fig:r50_STL10}, we select the MoCov2 encoder based on the ResNet50 backbone from the solo-learn repository to evaluate the performance of our proposed method on STL10 and GTSRB datasets.
STA and SRA represent the test accuracy and robust accuracy of the model under standard training, respectively.
The results indicate that Gen-AF also demonstrates strong generalizability and high robustness on the ResNet50 backbone.

\noindent\textbf{The effect of dataset size.} 
We explore the effect of downstream dataset sample size on our method. We randomly select subsets of $500$, $1000$, $2000$, and $5000$ samples from the STL10 dataset for training with Gen-AF. The results in the \cref{fig:ab_dataset_nums} show that as the number of training samples increases, both the accuracy and robustness of the model improve. This indicates that a sufficient downstream dataset aids in training a more robust and accurate downstream model.
\begin{table*}[t]
  \centering
  \caption{The ASR (\%) of different UAP schemes against downstream models under different settings. $\mathcal{D}_{1}$ - $\mathcal{D}_{6}$ denote the settings where the downstream datasets are Animals10, CIFAR10, GTSRB,  ImageNet20, STL10, and SVHN, respectively, and all the attacker’s surrogate dataset is CIFAR10.}    \scalebox{0.73}{
    \begin{tabular}{ccccccccccccc}
    \toprule[1.5pt]
     \rowcolor[gray]{0.9} \textbf{P-Dataset} & \textbf{Method}  & \textbf{Dataset} & \textbf{BYOL} & \textbf{DINO} & \textbf{MoCo2+} & \textbf{MoCo3} & \textbf{NNCLR} & \textbf{RESSL} & \textbf{SimCLR} & \textbf{SwAV} & \textbf{VibCreg} & \textbf{W-MSR} \\
\midrule
    \multirow{35}[10]{*}{\rotatebox{90}{\large CIFAR10}} & \multirow{7}[2]{*}{UAP} & $\mathcal{D}_{1}$ & 48.63 & 43.54 & 30.06 & 32.59 & 38.04 & 39.51 & 27.88 & 30.26 & 39.60 & \textbf{60.55} \\
          &       & $\mathcal{D}_{2}$ & 87.73 & \textbf{91.76} & 79.25 & 87.02 & 86.61 & 88.53 & 86.16 & 71.40 & 86.93 & 85.99 \\
          &       & $\mathcal{D}_{3}$ & 83.21 & 88.33 & 71.27 & 75.74 & 78.48 & 84.35 & 80.98 & 74.56 & 90.12 & \textbf{92.81} \\
          &       & $\mathcal{D}_{4}$ & 76.40 & 71.90 & 68.49 & 68.21 & 73.15 & 71.87 & 58.33 & 60.15 & 70.87 & \textbf{83.75} \\
          &       & $\mathcal{D}_{5}$ & 42.98 & 40.09 & 33.98 & 39.95 & 41.58 & 46.01 & 27.95 & 28.72 & 42.98 & \textbf{67.08} \\
          &       & $\mathcal{D}_{6}$ & 89.21 & 88.76 & 78.52 & 87.22 & 87.93 & 84.05 & 85.43 & 71.83 & 87.08 & \textbf{89.55} \\
          &       & AVG   & 71.36 & 70.73 & 60.26 & 65.12 & 67.63 & 69.05 & 61.12 & 56.15 & 69.60 & \textbf{79.96} \\
\cmidrule{2-13}          & \multirow{7}[2]{*}{UAPGD} & $\mathcal{D}_{1}$ & \textbf{56.81} & 41.57 & 17.16 & 14.68 & 39.23 & 36.60 & 17.01 & 18.35 & 39.16 & 20.61 \\
          &       & $\mathcal{D}_{2}$ & \textbf{89.35} & 89.02 & 22.67 & 25.09 & 89.14 & 88.47 & 13.13 & 23.15 & 87.65 & 37.86 \\
          &       & $\mathcal{D}_{3}$ & 87.40 & \textbf{89.28} & 45.74 & 40.90 & 79.13 & 84.29 & 35.21 & 40.70 & 89.21 & 51.48 \\
          &       & $\mathcal{D}_{4}$ & \textbf{83.69} & 68.36 & 49.51 & 45.56 & 73.27 & 70.20 & 40.85 & 43.96 & 70.87 & 46.74 \\
          &       & $\mathcal{D}_{5}$ & \textbf{53.94} & 38.96 & 16.61 & 12.62 & 46.12 & 42.70 & 11.41 & 15.60 & 42.61 & 18.52 \\
          &       & $\mathcal{D}_{6}$ & 89.94 & 87.45 & 23.02 & 25.19 & \textbf{90.25} & 85.95 & 13.44 & 21.78 & 84.40 & 40.04 \\
          &       & AVG   & \textbf{76.86} & 69.11 & 29.12 & 27.34 & 69.52 & 68.04 & 21.84 & 27.26 & 68.98 & 35.88 \\
\cmidrule{2-13}          & \multirow{7}[2]{*}{SSP} & $\mathcal{D}_{1}$ & 80.96 & 67.82 & 27.92 & 22.42 & 80.89 & 79.99 & 38.01 & \textbf{82.42} & 79.64 & 44.41 \\
          &       & $\mathcal{D}_{2}$ & 86.69 & \textbf{92.17} & 43.72 & 45.53 & 87.94 & 89.98 & 33.83 & 86.89 & 91.33 & 75.34 \\
          &       & $\mathcal{D}_{3}$ & 92.97 & 92.48 & 61.58 & 55.81 & 94.07 & 95.04 & 48.16 & 86.59 & \textbf{97.18} & 87.12 \\
          &       & $\mathcal{D}_{4}$ & 98.28 & 93.89 & 65.27 & 61.70 & 97.91 & \textbf{98.87} & 69.51 & 93.67 & 97.88 & 74.78 \\
          &       & $\mathcal{D}_{5}$ & 83.67 & 80.26 & 25.55 & 21.66 & 82.65 & \textbf{92.33} & 25.43 & 65.75 & 81.24 & 42.98 \\
          &       & $\mathcal{D}_{6}$ & 87.99 & 88.90 & 43.64 & 45.42 & 88.68 & \textbf{92.83} & 34.79 & 87.54 & 88.88 & 75.91 \\
          &       & AVG   & 88.43 & 85.92 & 44.61 & 42.09 & 88.69 & \textbf{91.51} & 41.62 & 83.81 & 89.36 & 66.76 \\
\cmidrule{2-13}          & \multirow{7}[2]{*}{PAP} & $\mathcal{D}_{1}$ & 81.89 & 77.03 & 42.78 & 36.41 & \textbf{86.86} & 79.99 & 53.12 & 41.20 & 81.21 & 75.43 \\
          &       & $\mathcal{D}_{2}$ & 89.64 & \textbf{92.19} & 56.75 & 77.55 & 88.93 & 89.98 & 55.95 & 71.92 & 90.12 & 86.04 \\
          &       & $\mathcal{D}_{3}$ & 93.31 & 94.88 & 76.12 & 69.22 & 96.82 & \textbf{97.89} & 70.83 & 66.10 & 94.76 & 90.52 \\
          &       & $\mathcal{D}_{4}$ & 98.58 & 98.16 & 80.75 & 73.98 & 98.45 & \textbf{98.90} & 84.52 & 70.89 & 97.36 & 93.35 \\
          &       & $\mathcal{D}_{5}$ & 84.89 & 85.31 & 41.21 & 40.89 & 85.78 & \textbf{89.82} & 39.70 & 35.86 & 86.55 & 71.48 \\
          &       & $\mathcal{D}_{6}$ & 91.10 & 88.79 & 58.06 & 76.98 & 89.14 & \textbf{92.18} & 56.93 & 69.22 & 89.59 & 84.11 \\
          &       & AVG   & 89.90 & 89.39 & 59.28 & 62.51 & 91.00 & \textbf{91.46} & 60.18 & 59.20 & 89.93 & 83.49 \\
\cmidrule{2-13}          & \multirow{7}[2]{*}{AdvEncoder} & $\mathcal{D}_{1}$ & \textbf{55.32} & 51.59 & 36.33 & 47.54 & 50.63 & 46.67 & 34.43 & 47.67 & 39.68 & 48.67 \\
          &       & $\mathcal{D}_{2}$ & 89.25 & \textbf{89.91} & 63.56 & 84.68 & 87.78 & 88.02 & 56.97 & 87.99 & 87.00 & 89.30 \\
          &       & $\mathcal{D}_{3}$ & 90.71 & 90.59 & 79.54 & \textbf{92.63} & 91.20 & 89.16 & 73.14 & 89.61 & 85.50 & 88.28 \\
          &       & $\mathcal{D}_{4}$ & \textbf{87.80} & 79.38 & 68.80 & 83.73 & 80.80 & 78.10 & 66.15 & 73.32 & 74.37 & 84.45 \\
          &       & $\mathcal{D}_{5}$ & \textbf{62.93} & 54.21 & 32.13 & 55.69 & 50.71 & 52.60 & 29.91 & 45.61 & 52.87 & 55.65 \\
          &       & $\mathcal{D}_{6}$ & \textbf{90.72} & 89.04 & 61.76 & 84.86 & 88.78 & 85.53 & 58.88 & 88.24 & 88.37 & 88.93 \\
          &       & AVG   & \textbf{79.46} & 75.79 & 57.02 & 74.86 & 74.98 & 73.35 & 53.25 & 72.07 & 71.30 & 75.88 \\
    \midrule
    \multirow{35}[10]{*}{\rotatebox{90}{\large ImageNet}} & \multirow{7}[2]{*}{UAP} & $\mathcal{D}_{1}$ & 45.57 & 40.84 & 44.25 & 39.83 & 42.82 & 44.87 & 45.62 & 45.11 & 41.20 & \textbf{51.06} \\
          &       & $\mathcal{D}_{2}$ & 80.35 & 76.35 & 80.57 & 68.71 & 76.76 & \textbf{86.71} & 81.17 & 82.01 & 66.23 & 85.64 \\
          &       & $\mathcal{D}_{3}$ & 68.00 & 67.46 & 74.14 & 64.46 & 70.00 & 77.08 & \textbf{77.27} & 74.14 & 59.91 & 77.23 \\
          &       & $\mathcal{D}_{4}$ & 64.99 & 67.62 & 65.50 & 63.75 & 63.50 & 66.44 & 67.37 & 66.78 & 65.79 & \textbf{79.52} \\
          &       & $\mathcal{D}_{5}$ & 44.26 & 49.49 & 46.03 & 47.23 & 46.77 & 45.70 & 53.26 & 42.60 & 43.62 & \textbf{60.01} \\
          &       & $\mathcal{D}_{6}$ & 75.79 & 67.24 & \textbf{82.08} & 66.33 & 72.48 & 81.81 & 78.67 & 76.70 & 65.45 & 81.46 \\
          &       & AVG   & 63.16 & 61.50 & 65.43 & 58.39 & 62.06 & 67.10 & 67.23 & 64.56 & 57.03 & \textbf{72.49} \\
\cmidrule{2-13}          & \multirow{7}[2]{*}{UAPGD} & $\mathcal{D}_{1}$ & 23.52 & 16.11 & 18.50 & 18.40 & 20.09 & 19.23 & 23.66 & 22.15 & 17.18 & \textbf{25.97} \\
          &       & $\mathcal{D}_{2}$ & 34.89 & 29.56 & 31.41 & 32.37 & 29.70 & 32.55 & 36.36 & 35.05 & 25.23 & \textbf{46.68} \\
          &       & $\mathcal{D}_{3}$ & 35.24 & 31.59 & 30.31 & 34.53 & 31.18 & 31.28 & 31.02 & 30.76 & 23.29 & \textbf{48.09} \\
          &       & $\mathcal{D}_{4}$ & 41.05 & 33.44 & 33.47 & 32.51 & 32.33 & 34.26 & 37.59 & 38.94 & 31.36 & \textbf{44.94} \\
          &       & $\mathcal{D}_{5}$ & 22.69 & 16.70 & 19.37 & 19.18 & 17.72 & 18.15 & 20.79 & 19.43 & 14.16 & \textbf{28.00} \\
          &       & $\mathcal{D}_{6}$ & 34.73 & 27.19 & 30.37 & 32.28 & 28.06 & 31.77 & 36.00 & 33.15 & 26.14 & \textbf{49.17} \\
          &       & AVG   & 32.02 & 25.77 & 27.24 & 28.21 & 26.51 & 27.87 & 30.90 & 29.91 & 22.89 & \textbf{40.48} \\
\cmidrule{2-13}          & \multirow{7}[2]{*}{SSP} & $\mathcal{D}_{1}$ & 42.82 & 34.84 & 44.69 & 41.15 & 42.89 & \textbf{58.52} & 50.13 & 50.88 & 40.24 & 46.89 \\
          &       & $\mathcal{D}_{2}$ & 79.04 & 60.96 & 71.72 & 75.05 & 71.16 & \textbf{89.38} & 81.02 & 84.27 & 73.01 & 77.96 \\
          &       & $\mathcal{D}_{3}$ & 76.72 & 70.59 & 74.82 & 79.24 & 69.45 & \textbf{83.27} & 80.72 & 78.19 & 72.87 & 76.95 \\
          &       & $\mathcal{D}_{4}$ & 63.15 & 60.63 & 64.30 & 64.64 & 62.80 & \textbf{76.69} & 67.72 & 72.57 & 65.99 & 66.66 \\
          &       & $\mathcal{D}_{5}$ & 47.80 & 41.70 & 49.18 & 51.86 & 51.17 & \textbf{61.94} & 50.12 & 44.63 & 49.73 & 44.98 \\
          &       & $\mathcal{D}_{6}$ & 76.12 & 60.81 & 71.65 & 76.21 & 67.04 & \textbf{84.63} & 78.87 & 78.31 & 70.91 & 80.29 \\
          &       & AVG   & 64.28 & 54.92 & 62.73 & 64.69 & 60.75 & \textbf{75.74} & 68.10 & 68.14 & 62.13 & 65.62 \\
\cmidrule{2-13}          & \multirow{7}[2]{*}{PAP} & $\mathcal{D}_{1}$ & 37.25 & 33.92 & 37.43 & 35.41 & 39.04 & \textbf{63.90} & 50.55 & 38.51 & 37.11 & 56.82 \\
          &       & $\mathcal{D}_{2}$ & 53.87 & 48.04 & 42.96 & 49.84 & 56.87 & \textbf{89.57} & 77.94 & 55.88 & 50.75 & 80.45 \\
          &       & $\mathcal{D}_{3}$ & 60.59 & 55.24 & 53.28 & 62.20 & 59.07 & \textbf{83.80} & 74.44 & 60.39 & 53.95 & 68.71 \\
          &       & $\mathcal{D}_{4}$ & 57.63 & 55.04 & 54.89 & 55.34 & 57.93 & 80.09 & 69.97 & 61.08 & 58.26 & \textbf{82.76} \\
          &       & $\mathcal{D}_{5}$ & 37.77 & 29.61 & 36.62 & 39.01 & 41.67 & \textbf{66.52} & 56.24 & 36.45 & 39.90 & 61.85 \\
          &       & $\mathcal{D}_{6}$ & 52.48 & 44.29 & 42.19 & 51.93 & 54.86 & \textbf{84.67} & 78.58 & 55.29 & 51.44 & 79.79 \\
          &       & AVG   & 49.93 & 44.36 & 44.56 & 48.96 & 51.57 & \textbf{78.09} & 67.95 & 51.27 & 48.57 & 71.73 \\
\cmidrule{2-13}          & \multirow{7}[2]{*}{AdvEncoder} & $\mathcal{D}_{1}$ & 53.72 & 47.66 & 47.33 & 46.45 & 46.65 & 56.28 & 49.83 & 49.78 & 48.52 & \textbf{57.70} \\
          &       & $\mathcal{D}_{2}$ & 85.59 & 84.19 & 84.07 & 73.24 & 71.45 & 88.94 & 70.75 & 84.88 & 81.24 & \textbf{89.12} \\
          &       & $\mathcal{D}_{3}$ & 76.66 & 73.70 & 78.14 & 70.79 & 72.96 & 84.45 & 66.32 & 71.56 & 75.75 & \textbf{85.17} \\
          &       & $\mathcal{D}_{4}$ & 75.58 & 77.80 & 74.78 & 65.49 & 69.26 & 75.79 & 71.04 & 74.15 & 72.60 & \textbf{84.05} \\
          &       & $\mathcal{D}_{5}$ & 56.16 & 61.62 & 53.52 & 59.88 & 52.76 & 59.24 & 55.24 & 55.94 & 58.81 & \textbf{67.44} \\
          &       & $\mathcal{D}_{6}$ & 83.56 & 74.82 & \textbf{86.93} & 73.25 & 66.08 & 83.74 & 71.01 & 83.77 & 77.80 & 83.73 \\
          &       & AVG   & 71.88 & 69.97 & 70.80 & 64.85 & 63.19 & 74.74 & 64.03 & 70.01 & 69.12 & \textbf{77.87} \\
    \bottomrule[1.5pt]
\end{tabular}%
    }
  \label{tab:attack}%
\end{table*}%

\end{document}